\documentclass[preprint,12pt]{elsarticle}

\usepackage{url}
\usepackage{cite}
\usepackage{amsmath,amssymb,amsfonts}
\usepackage{algorithmic}
\usepackage{graphicx}
\usepackage{textcomp}
\usepackage{url}
\usepackage{booktabs}
\usepackage{stfloats}
\usepackage{tabularx}
\usepackage{multirow}
\usepackage{cleveref}
\usepackage{mathtools}
\usepackage{subcaption}
\usepackage{xcolor}

\definecolor{modicolor}{rgb}{0,0,0}

\definecolor{modicolor2}{rgb}{0,0,0}

\journal{Medical Image Analysis}

\begin{document}

\def\modelName{{PASTA}}

\newcommand{\interval}[1]{\scriptsize{$\pm$ #1}}
\newcommand{\intervalsmall}[1]{\tiny{$\pm$ #1}}

\begin{frontmatter}



\title{Translating MRI to PET through Conditional Diffusion Models with Enhanced Pathology Awareness} 



\author[label1,label2]{Yitong Li\corref{cor1}}
\cortext[cor1]{Corresponding author}
\ead{yi_tong.li@tum.de}
\author[label3]{Igor Yakushev}
\ead{igor.yakushev@tum.de}
\author[label4]{Dennis M. Hedderich}
\ead{dennis.hedderich@tum.de}
\author[label1,label2]{Christian Wachinger}
\ead{christian.wachinger@tum.de}
\affiliation[label1]{organization={Lab for Artificial Intelligence in Medical Imaging, Institute for Diagnostic and Interventional Radiology, School of Medicine and Health, TUM Klinikum, Technical University of Munich (TUM)},
            city={Munich},
            postcode={81675},
            country={Germany}}

\affiliation[label2]{organization={Munich Center for Machine Learning (MCML)},
            city={Munich},
            country={Germany}}
\affiliation[label3]{organization={Department of Nuclear Medicine, School of Medicine and Health},
            addressline={TUM Klinikum},
            city={Munich},
            postcode={81675},
            country={Germany}}
\affiliation[label4]{organization={Department of Neuroradiology, School of Medicine and Health},
            addressline={TUM Klinikum},
            city={Munich},
            postcode={81675},
            country={Germany}}



\begin{abstract}

Positron emission tomography (PET) is a widely recognized technique for diagnosing neurodegenerative diseases, offering critical functional insights.
However, its high costs and radiation exposure hinder its widespread use. In contrast, magnetic resonance imaging (MRI) does not involve such limitations. While MRI also detects neurodegenerative changes, it is less sensitive for diagnosis compared to PET. 
To overcome such limitations, one approach is to generate synthetic PET from MRI.
Recent advances in generative models have paved the way for cross-modality medical image translation; however, existing methods largely emphasize structural preservation while neglecting the critical need for pathology awareness.
To address this gap, we propose PASTA, a novel image translation framework built on conditional diffusion models with enhanced pathology awareness. \modelName~surpasses  state-of-the-art methods by preserving both structural and pathological details through its highly interactive dual-arm architecture and multi-modal condition integration. 
Additionally, we introduce a novel cycle exchange consistency and volumetric generation strategy that significantly enhances PASTA's ability to produce high-quality 3D PET images.
Our qualitative and quantitative results demonstrate the high quality and pathology awareness of the synthesized PET scans. For Alzheimer's diagnosis, the performance of these synthesized scans improves over MRI by 4\%, almost reaching the performance of actual PET. Our code is available at~\url{https://github.com/ai-med/PASTA}.
\end{abstract}



\begin{keyword}



Diffusion models \sep cross-modality translation \sep brain \sep  MRI \sep PET.

\end{keyword}

\end{frontmatter}



\section{Introduction}
\label{sec:intro}

Alzheimer's disease (AD) is a progressive neurodegenerative disorder and the leading cause of dementia; early detection is vital for timely therapeutic interventions. 
Accurately diagnosing a patient’s neurological condition requires multidisciplinary diagnostic tools, including magnetic resonance imaging (MRI), positron emission tomography (PET), cognitive assessments, and genetic tests~\citep{addiagnosis}. 
Structural MRI provides detailed anatomical information about the brain, facilitating the identification of regional atrophy, a hallmark of AD~\citep{mri_pet}. 
In contrast, PET with 18-Fluorodeoxyglucose (FDG$^{18}$-PET) measures glucose metabolism in the brain.
In AD and other neurodegenerative disorders, glucose uptake is severely reduced in specific brain regions~\citep{ad_glucose}. By sensitively capturing functional abnormalities, PET is highly effective for assessing early dementia symptoms and distinguishing AD from other types of dementia, such as frontotemporal and Lewy body dementia~\citep{pet_differential}. As a result, PET is widely regarded as having a higher diagnostic and prognostic accuracy for AD~\citep{pet_acc}.

Despite its considerable clinical value, PET is inaccessible to numerous medical centers worldwide due to its high cost and the inherent risks associated with radiation exposure~\citep{pet_cost}. MRI, while being more widely available and non-invasive, lacks the functional insights provided by PET, making it less sensitive for such diagnostic purposes~\citep{pet_ov_mri}. 
To address this limitation, one promising approach is to translate MRI data into synthetic PET images, thereby improving access to functional brain imaging and facilitating early, accurate AD diagnosis. However, a major challenge in such cross-modality translation is faithfully preserving and generating accurate pathological features in the target imaging modality. Introducing artificial or incorrect pathology will compromise the reliability of synthetic images, rendering them unsuitable for clinical applications and increasing the risk of misdiagnosis. Therefore, ensuring \textit{pathology awareness} in cross-modality translation is essential for its safe and effective use in healthcare. 

\begin{figure}[t]
    \centering
    \includegraphics[width=\linewidth]{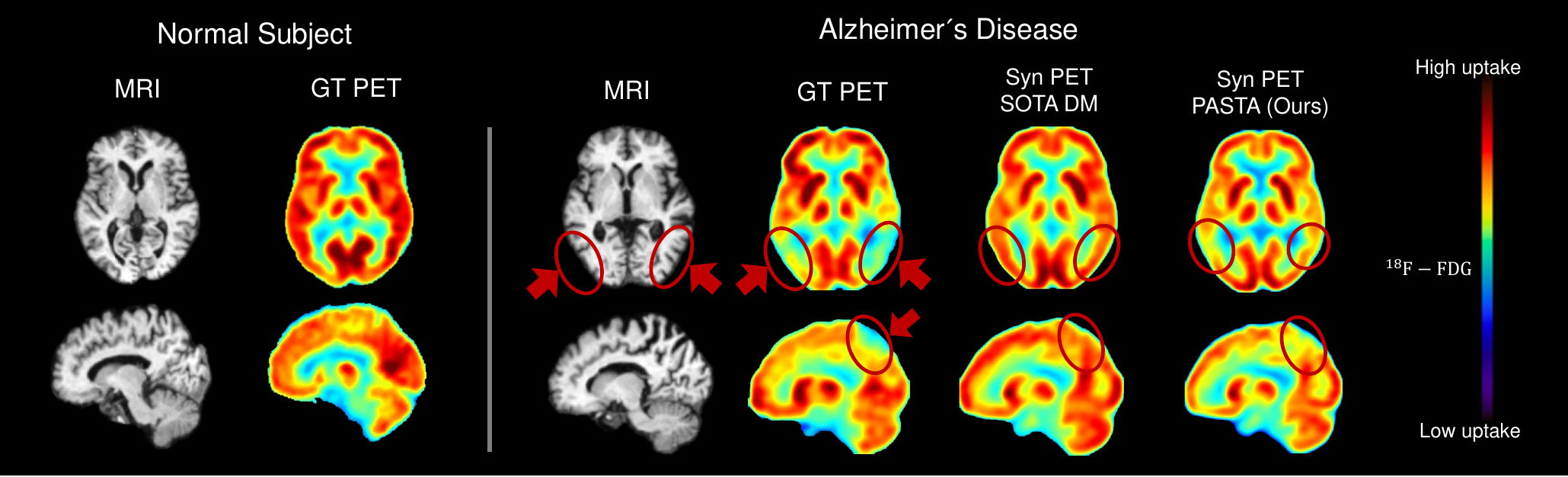}
    \caption{For Alzheimer's disease, PET reveals distinctly reduced glucose uptake in the temporoparietal lobe (bottom circles), mirroring atrophy on MRI with higher sensitivity. Compared to the ground-truth (GT) PET, state-of-the-art diffusion models (SOTA DM) fail to recover such pathology in the synthesized (Syn) PET from MRI input. In contrast, PASTA shows improvement in preserving disease-specific pathology.}
    \label{fig:intro}
\end{figure}

Recently, generative models, particularly diffusion models (DM)~\citep{ddpm}, have gained significant attention for their exceptional capabilities in high-quality image generation and translation~\citep{DMbeatsGAN, bbdm, Palette}, making them a compelling choice for cross-modality medical imaging translation.
However, current DM-based approaches predominantly emphasize maintaining structural integrity, often neglecting the preservation of pathology, as shown in Fig.~\ref{fig:intro}.

To address this problem, we present Pathology-Aware croSs-modal TrAnslation (PASTA), a novel end-to-end DM-based framework for clinically meaningful volumetric MRI to PET translation.
It is based on a symmetric dual-arm architecture consisting of a conditioner arm, a denoiser arm, and intermediate adaptive conditional modules to integrate multi-modal conditions. The conditioner arm processes the MRI input and generates a task-specific representation, passed on to condition the denoiser arm through adaptive conditional modules, together with the provided clinical data. The denoiser arm utilizes the conditions to generate corresponding PET scans from complete noises.
In addition, \modelName~introduces a memory-efficient volumetric generation paradigm, which leverages 2D backbones to create high-quality scans without any inconsistencies or artifacts in the 3D representation. 
Finally, we introduce a cycle exchange consistency training strategy for PASTA to enforce the information exchange within the dual-arm architecture, further lifting the generation quality. As illustrated in Fig.~\ref{fig:intro}, PASTA represents a significant step forward in supporting AD diagnosis through high-quality synthetic PET imaging.


In summary, we make the following contributions:

\begin{itemize}
    \item A novel end-to-end framework for cross-modality MRI to PET translation with pathology-aware conditional diffusion models for volumetric image generation.
    \item Integration of multi-modal conditions through adaptive normalization layers to facilitate high-quality PET synthesis enabling pathology awareness.
    \item A cycle exchange consistency strategy for informative training of conditional diffusion models.
    \item Quantitative and qualitative experiments demonstrate that PASTA not only achieves low reconstruction errors but also preserves AD pathology to boost diagnosis accuracy.
\end{itemize}
%


\section{Related Work}
\label{sec:related_works}

\subsection{Image Translation in Medical Domains}
Previous works on cross-modality MRI to PET translation mainly based on generative adversarial networks (GAN)~\citep{sketcher-refine,SCGAN,bpgan,bidirectionalGAN,ganbert,resvit,gandalf}. Wei \textit{et al.}~\citep{sketcher-refine} proposed a sketcher-refiner scheme with two cascaded GANs to first generate coarse synthetic images and then refine them. Shin \textit{et al.}~\citep{gandalf} introduced GANDALF to generate PET from MRI for Alzheimer's diagnosis. Zhang \textit{et al.}~\citep{bpgan} proposed BPGAN to synthesize brain PET images. Hu \textit{et al.}~\citep{bidirectionalGAN} designed a 3D end-to-end synthesis model called bidirectional mapping GAN, in which the image context and the latent vector were jointly optimized. Dalmaz \textit{et al.}~\citep{resvit} proposed a GAN-based residual vision Transformers framework for multi-modal medical image synthesis. 
\textcolor{modicolor}{RegGAN was introduced by Kong \textit{et al.}~\citep{kong2021breaking} for unpaired medical image-to-image translation between MR T1- and T2-weighted images, with an additional registration network to fit the misaligned noise distribution adaptively.
Lin \textit{et al.}~\citep{lin2021bidirectional} adopted a 3D Reversible GAN (RevGAN) to synthesize missing PET from MRI, so that incomplete multimodal data could be fully utilized in a 3D CNN classification model to perform AD diagnosis. 
A Plasma-CycleGAN was proposed by Chen \textit{et al.}~\citep{chen2025plasma} to synthesize PET images from MRI using blood-based biomarkers. In an epilepsy study conducted by Zotova \textit{et al.}~\citep{zotova2025gan}, training an unsupervised anomaly detection model on synthetic FDG-PET scans from GAN-based models achieved lesion detection performance on par with using real PET.}

\subsection{Diffusion Models}
Diffusion models (DM) excel at modeling complex distributions by leveraging parameterized Markov chains to optimize the lower variational bound on the likelihood function~\citep{diffusion, ddpm}. They have achieved superior performance over GANs in generation fidelity and diversity~\citep{DMbeatsGAN}. 
Recently, DMs have been widely adopted for medical imaging, which mainly focus on unconditional medical image synthesis or cross-contrast MRI translation~\citep{make-a-volume,generate_real_MRI_cDPM,syndiff}\textcolor{modicolor}{\citep{kim2024adaptive}}.
\textcolor{modicolor}{
A diffusion-based framework, Make-A-Volume, was presented by Zhu \textit{et al.}~\citep{make-a-volume} for cross-modality Brain MRI Synthesis, which extended the 2D latent diffusion model to a volumetric version with volumetric layers in the 2D slice-mapping model followed by fine-tuning with 3D data.
Peng \textit{et al.}~\citep{generate_real_MRI_cDPM} trained a 2D conditional Diffusion Probabilistic Model to generate unconditional realistic brain MRIs by progressively generating MRI slices based on previously generated slices.
Based on adversarial diffusion modeling, Özbey \textit{et al.}~\citep{syndiff} proposed SynDiff for multi-contrast MRI and MRI-CT translation in an unsupervised setup.
Kim \textit{et al.}~\citep{kim2024adaptive} proposed a latent diffusion model that leverages switchable blocks for multi-modal 3D MRI image-to-image translation.
}
Considering their remarkable achievements, harnessing DMs for cross-modality MRI to PET translation presents a promising avenue. 
\textcolor{modicolor}{Recent work by Xie \textit{et al.}~\citep{xie2024synpet} proposed a Joint Diffusion Attention Model (JDAM) to synthesize PET from high-field and ultra-high-field MR images. 
Yu \textit{et al.}~\citep{yu2024functional} introduced a Functional Imaging Constrained Diffusion (FICD) framework for brain PET image synthesis with paired structural MRI as input condition, through a new constrained diffusion model.
Chen \textit{et al.}~\citep{chen2025multi} proposed using a diffusion model to synthesize FDG-PET views from T1-weighted MRI views and incorporate both one-way and two-way synthesis strategies.
With such recent ongoing research, diffusion models are becoming a more powerful tool for generating synthetic PET with enhanced accuracy, complementing or even surpassing GAN-based approaches.
}
\\

A preliminary version of this work has been presented at a conference~\citep{Li2024pasta}. Here, we extend it by providing more technical details \textcolor{modicolor}{with new architectural improvements}, extending the experimental evaluation on more datasets, including more analysis on the Neurostat 3D-SSP maps from the generated images, adding analysis on the influence of individual clinical data on the generative performance, and including additional ablation studies with error maps to showcase the impact of various critical designs.


\section{Methods}
\label{sec:methods}

Cross-modality image translation aims to learn a mapping from one modality to another in a paired manner, given datasets $\mathcal{X}_{A}$ and $\mathcal{X}_{B}$ from modalities A and B, respectively. In medical image translation, it is crucial that the generated images closely follow the ground truth and preserve pathological features. Existing DM-based image translation methods~\citep{ddib,diffuseIT} focus on transferring image style and preserving structural information; however, these approaches are insufficient for medical image translation in the alignment of pathology details, as shown in Fig.~\ref{fig:intro}. 
PASTA addresses this limitation in translating 3D brain MRI scans to PET 
using conditional denoising diffusion probabilistic models (DDPM). In the following section, we first review the fundamental concepts of DDPM for data generation, then introduce \modelName~and its efficient strategy for generating pathology-aware 3D PET scans from corresponding MRIs.


\subsection{Preliminaries of DDPMs}
A $T$-step DDPM~\citep{ddpm} comprises a forward diffusion process and a reverse denoising process. Denoting the distribution of training data as $p(x_0)$, the diffusion process is a Markovian Gaussian transition that gradually adds noise with different scales to a real data point $x_0 \sim p(x_0)$ to obtain a series of noisy latent variables $ \{x_1, x_2, ..., x_T\} $:
\begin{equation}\tag{\textcolor{modicolor}{1}}
\begin{aligned}
    q(x_t | x_0) &= \mathcal{N} (x_t; \alpha_t x_0, \sigma_t^2 \mathbf{I}), \\
    x_t &= \alpha_t x _0 + \sigma_t \epsilon, 
\end{aligned}
\end{equation}
where $\epsilon \in \mathcal{N}(0, \mathbf{I})$, $\sigma_t$ is the noise schedule denoting the magnitude of noise added to the original data at timestep $t$, increasing monotonically. We adopt the standard variance-preserving diffusion process, where $\alpha_t = \sqrt{1-\sigma_t^2}$.

The reverse process gradually denoises the latent variables and restores the clean data $x_0$ from $x_T$ by approximating the posterior distribution $p_\theta (x_{t-1} \mid x_t)$, parameterized as a  Gaussian transition. The denoising process goes through the entire Markov chain from timestep $T$ to $0$, given by:
\begin{align}
    p_\theta (x_{0:T}) &\coloneqq p(x_T)\prod_{t=1}^T  p_\theta (x_{t-1} \mid x_t), \tag{\textcolor{modicolor}{2}} \\
    p_\theta (x_{t-1} \mid x_t) &\coloneqq \mathcal{N}(x_{t-1}; \hat{\mu}_\theta(x_t, t), \hat{\Sigma}_\theta(x_t, t)), \tag{\textcolor{modicolor}{3}}
\end{align}
where $\hat{\mu}$ and $\hat{\Sigma}$ are both predicted statistics. Ho et al.~\citep{ddpm} find that instead of learning $\hat{\Sigma} (x_t, t)$, they can fix it to a constant $\sigma_t^2 \mathbf{I}$ or $\Tilde{\sigma_t}^2 \mathbf{I}$, corresponding to upper or lower bounds for the true reverse step variance. $\hat{\mu_\theta}$ can be decomposed into the linear combination of $x_t$ and a noise approximation model $\hat{\boldsymbol{\epsilon}_\theta}$. They find that instead of directly parameterizing $\mu_\theta(x_t, t)$ as a neural network, using a model $\boldsymbol{\epsilon}_\theta(x_t, t)$ to predict input noise $\epsilon$ yields better performance in practice, \textcolor{modicolor}{leading to} a simplified objective:
\begin{equation}
    \mathcal{L}_{simple}^t(\theta) = \mathbb{E}_{x_0, \epsilon} [\| \epsilon - \hat{\boldsymbol{\epsilon}}_\theta (\alpha_t x_0 + \sigma_t \epsilon) \|_2^2]. \tag{\textcolor{modicolor}{4}}
\end{equation}
Most works~\citep{improved_ddpm, DMbeatsGAN} adopt this strategy. Later works~\citep{stable_diffusion, progressive} also use another reparameterization that trains the denoising model $\mathbf{x}_{\theta}(x_t, t)$ to predict the noiseless state $x_0$ as it gives better empirical results for specific models:
\begin{equation}
\label{equ:general_loss}
    \mathcal{L}_{simple}^t(\theta) = \mathbb{E}_{x_0, \epsilon} [\| x_0 - \hat{\mathbf{x}}_\theta (\alpha_t x_0 + \sigma_t \epsilon) \|_2^2]. \tag{\textcolor{modicolor}{5}}
\end{equation}
Despite their different prediction targets, these objectives are mathematically equivalent~\citep{stable_diffusion}. In this paper, we stick to the strategy of training the denoising model $\mathbf{x}_\theta$ with the objective in \cref{equ:general_loss}, for its empirically better performance.

\subsection{Conditional PET Generation from MRI}

\begin{figure}[t]
  \centering
  \includegraphics[width=\textwidth]{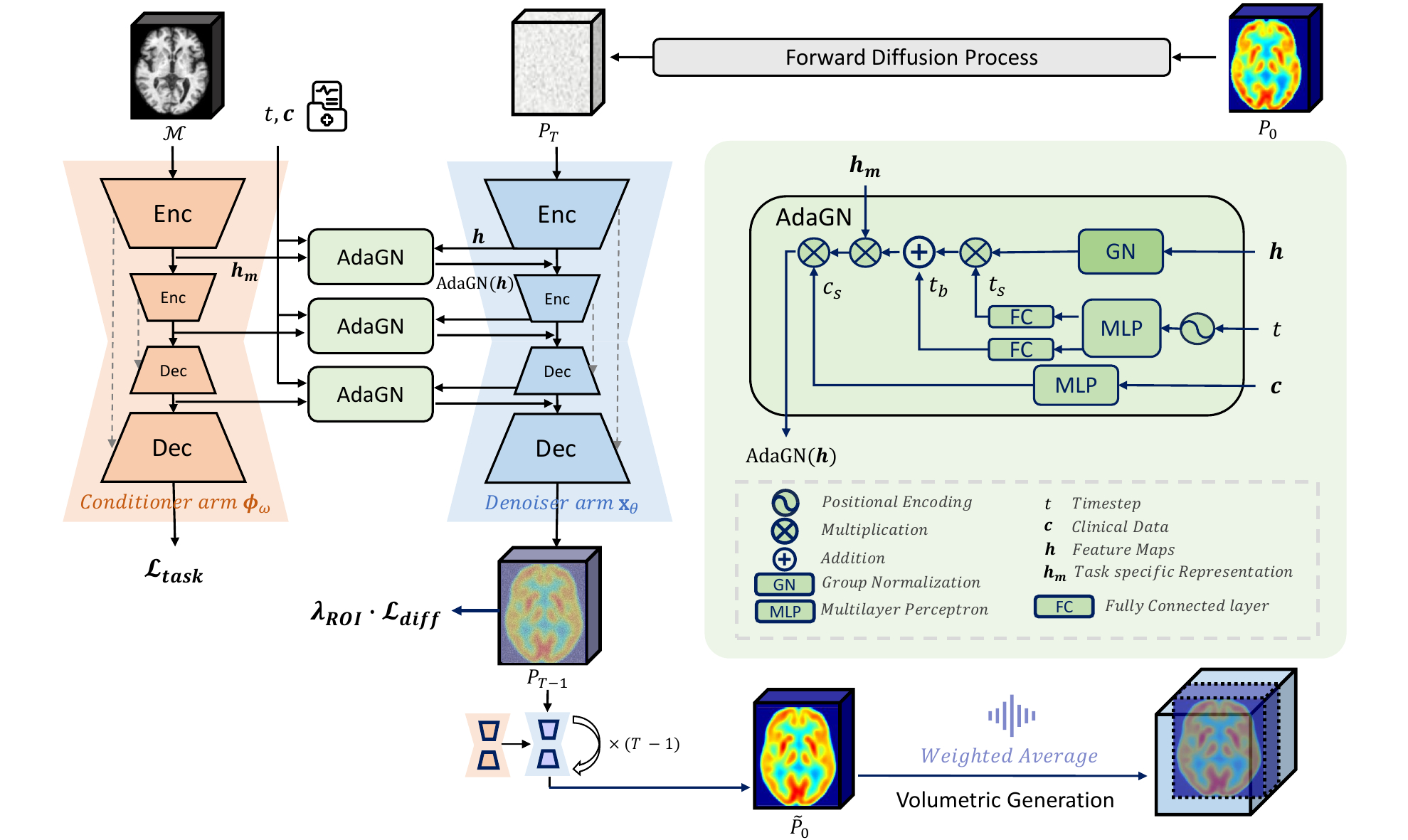}
  \caption{PASTA features a symmetric dual-arm structure with a conditioner arm ($\phi_\omega$), a denoiser arm ($\mathbf{x}_\theta$), and adaptive conditional modules (AdaGN). Through AdaGN, \modelName~conditions the feature maps $\boldsymbol{h}$ from $\mathbf{x}_\theta$ on timestep $t$, clinical data $\boldsymbol{c}$, and task representation $\boldsymbol{h}_m$ from $\phi_\omega$. It achieves high-quality 3D PET synthesis through a volumetric generation strategy.}
  \label{fig:architecture}
\end{figure}

Denoting the training data as $\mathcal{D}_{\text{Train}} = {(\mathcal{M}^i_{\text{Train}}, \mathcal{P}^i_{\text{Train}})}_{i=1}^N$, which comprises $N$ pairs of MRI $\mathcal{M} \in \mathbb{R}^{H \times W \times D}$ and its corresponding PET $\mathcal{P} \in \mathbb{R}^{H \times W \times D}$, our objective is to learn a model $\boldsymbol{G}(\cdot)$ on $\mathcal{D}_{\text{Train}}$, such that given any unseen MRI input $\mathcal{M}_{\text{Test}} \notin \mathcal{D}_{\text{Train}}$, its PET counterpart is inferred as $\mathcal{P}_{\text{Test}} = \boldsymbol{G}(\mathcal{M}_{\text{Test}})$.
We propose a conditional DDPM to model $\boldsymbol{G}(\cdot)$, which is tailored to closely capture the inherent structural and pathological correlations present in the MRI data, with the integration of multi-modal conditions, to strive for a pathology-aware MRI to PET translation. We aim to establish a strong interaction with the input MRI by conditioning on its features at multiple scales. 
Thus, \modelName~deploys a symmetric dual-arm architecture, as shown in Fig.~\ref{fig:architecture}. The framework consists of three main parts: the conditioner arm, the denoiser arm, and the intermediate adaptive conditional modules. Both the conditioner and denoiser arms adopt a symmetric UNet~\citep{unet} layout. The interaction between the two arms and the fusion of additional conditions are achieved through the adaptive group normalization layers (AdaGN)~\citep{DMbeatsGAN}. This symmetric design ensures that matching blocks across the two arms share the same spatial resolution, enabling multi-scale feature map interactions. 

\subsubsection{Conditioner Arm}

The MRI input $\mathcal{M}$ is first processed through the conditioner arm to generate multi-scale task-specific representations, denoted as $\boldsymbol{h}_m$:
\begin{equation}
 \hat{\mathcal{M}} = \phi_\omega(\mathcal{M}; \boldsymbol{h}_m), \tag{\textcolor{modicolor}{6}} \\
\end{equation}
where $\phi_\omega(\cdot)$ represents the conditioner model parameterized by $\omega$, the task-specific representations $\boldsymbol{h}_m = \{\boldsymbol{h}_m^1, \dots, \boldsymbol{h}_m^n\}$ consist of intermediate feature maps from $\phi_\omega(\cdot)$ at multiple scales, with $n$ denoting the number of residual blocks in the conditioner arm. These representations play a key role in facilitating the PET synthesis procedure in the other arm. 

The conditioner arm is designed to perform predefined tasks to transform the MRI input into meaningful multi-scale features. These predefined tasks include MRI reconstruction, MRI-to-PET translation. Different tasks impose distinct training objectives on the conditioner arm. 
For instance, when MRI-to-PET translation is selected as the predefined task, the conditioner arm is trained by minimizing the pixel-wise distance between the original PET and the conditioner output:
\begin{equation}
    \mathcal{L}_{task}(\omega) = \mathbb{E}_{\mathcal{M}, \mathcal{P}}~dist(\phi_\omega(\mathcal{M}), \mathcal{P}), \tag{\textcolor{modicolor}{7}}
\end{equation}
where $dist(\cdot)$ denotes the distance function used to measure the similarity between $\phi_\omega(\mathcal{M})$ and $\mathcal{P}$, such as $L_1$ or $L_2$ norm.
MRI-to-PET translation is primarily chosen as the predefined task for the conditioner arm due to its superior empirical performance. Additionally, we explore the efficacy of alternative task paradigms in \Cref{sec:experiments}.

\subsubsection{Adaptive Conditional Module}
The interaction between the two arms is facilitated by the adaptive group normalization layers (AdaGN)~\citep{DMbeatsGAN}. These layers integrate multi-modal conditions into the feature maps $\boldsymbol{h} = \{\boldsymbol{h}^1, \dots, \boldsymbol{h}^n\}$ within each residual block from the denoiser arm. These AdaGN layers adapt conditions include: 1) Timestep $t$ in the diffusion process; 2) Task-specific representations $\boldsymbol{h}_m$ from the conditioner arm at corresponding scales; 3) Clinical data $\boldsymbol{c} \in \mathbb{R}^{c \times n}$ of an individual subject. Our AdaGN is given by:
\begin{align}
\label{eq:adagn}
    \textrm {AdaGN}&(\boldsymbol{h}, t, \boldsymbol{c}, \boldsymbol{h_m}) = \boldsymbol{c_s}(\boldsymbol{h_m} (\boldsymbol{t_s}\textrm{GroupNorm}(\boldsymbol{h})+\boldsymbol{t_b}), \tag{\textcolor{modicolor}{8}}
\end{align}
 in which $(\boldsymbol{t}_s, \boldsymbol{t}_b) \in \mathbb{R}^{2\times c} = \textrm{MLP}(pos(t))$ is the output of a multilayer perceptron (MLP) with a sinusoidal encoding function $pos(\cdot)$ applied to timestep $t$, and $\boldsymbol{c}_s = \textrm{MLP}(\boldsymbol{c})$. When incorporating the task-specific representations $\boldsymbol{h}_m$, we omit the linear projection but directly apply the feature maps after the timestep condition. 
 \textcolor{modicolor}{The adaptive conditional module can be further enhanced with slice-position awareness by incorporating through-plane positional information as an additional conditioning signal into AdaGN, in the same manner as the timestep and other conditions. This extension yields the Slice-Aware Adaptive Group Normalization (SA-AdaGN) module, which explicitly encodes the slice position into the conditioning process, aiming to improve the inter-slice consistency during generation. A detailed description of SA-AdaGN is provided in Appendix B.}
 These adaptive conditional modules are applied throughout the dual-arm architecture.

To enhance the network's capability to preserve detailed pathological evidence in the synthesized PET scans, we integrate clinical data as a supplementary condition during the translation process. 
As a default, we selected six variables for the clinical data $\boldsymbol{c}$: demographic variables (age, gender, education level), cognitive scores (MMSE~\citep{mmse}, ADAS-Cog-13~\citep{adas-cog-13}), and the AD-related genetic risk factor ApoE4~\citep{apoe4}. To account for variables that are not always available, we address missing values using a strategy inspired by Jarrett \textit{et al.}~\citep{tabularselect}, where binary indicators are appended to denote whether the data is missing or not for each clinical variable, except age, gender, and education level as they have no missing values. This allows the network to leverage incomplete data while recognizing patterns of missingness. The resulting clinical data includes nine features. 
  In this way, the adaptive conditional module fuses the multi-modal data ranging from structural to pathological evidence, enhancing the accuracy and reliability of the medical image synthesis in the following denoiser arm.

\subsubsection{Denoiser Arm}

The denoiser arm implements the reverse process of DDPM, aiming to restore the clean PET scan $\mathcal{P}_0$ from the noised input. Starting from the Gaussian noise $\epsilon$ at timestep $t = T$, it generates PET images $\mathcal{P}_t$ at each timestep $t = T, T-1, \dots, 0$, conditioned on multi-modal variables through AdaGN modules:
\begin{equation}
  \mathcal{P}_{0:T} = \mathcal{P}_T\prod_{t=1}^T  \mathbf{x}_\theta \left(\mathcal{P}_{t-1} \mid \boldsymbol{h}_m, \boldsymbol{c}\right), \tag{\textcolor{modicolor}{9}}
\end{equation}
where $\mathbf{x}_\theta(\cdot)$ denotes the denoising model parameterized by $\theta$. The symmetric layout of \modelName~enables feature maps in the denoiser arm at each respective scale to be conditioned by task-specific representations from the conditioner arm at corresponding scales. This design fosters stronger interactions with the conditioning modality, thereby strengthening its impact on the denoising process.

To further enhance the pathology awareness during the synthesis process, we incorporate MetaROIs~\citep{metaROIs} as pathology priors, guiding the model to focus on key hypometabolic regions associated with abnormal metabolic changes in AD patients. As concluded by Landau~\textit{et al.}, MetaROIs are a set of pre-defined regions of interest, derived from frequently cited coordinates in other PET studies comparing AD patients to normal subjects. The final MetaROIs consist of five regions: the left and right angular gyrus, the bilateral posterior cingular, and the left and right inferior temporal gyrus.
We transform the MetaROIs into a loss weighting map for the denoiser arm, denoted as $\boldsymbol{\lambda}_{R} \in \mathbb{R}^{H \times W \times D}$
\textcolor{modicolor2}{, with a constant weight factor $\lambda_{R}$ on the MetaROIs area}. 
During training, this weighting map emphasizes the MetaROIs within the denoised PET images, assigning higher penalties to deviations from the ground-truth PET in these critical regions.  The resulting training objective for the denoiser arm is:
\begin{equation}
    \mathcal{L}^t_{diff}(\theta) = \boldsymbol{\lambda}_{R}\cdot\mathbb{E}_{\mathcal{P}_0, \epsilon} [\| \mathcal{P}_0 - {\mathbf{x}}_\theta (\alpha_t \mathcal{P}_0 + \sigma_t \epsilon, \boldsymbol{c}, \boldsymbol{h}_m) \|_2^2]. \tag{\textcolor{modicolor}{10}}
\end{equation}
\textcolor{modicolor}{This training process can be further augmented with an in-loop auxiliary classifier that enforces disease-label consistency on the clean PET images predicted by the denoiser arm. By introducing this auxiliary supervision, we couple a discriminative signal directly into the generation process, guiding the synthesis toward pathology-aware and semantically faithful outputs while also offering potential benefits for interpretability. A detailed description of this auxiliary classifier consistency loss is provided in Appendix C.}

\subsubsection{Cycle Exchange Consistency}

\begin{figure}[t]
    \centering
    \includegraphics[width=\linewidth]{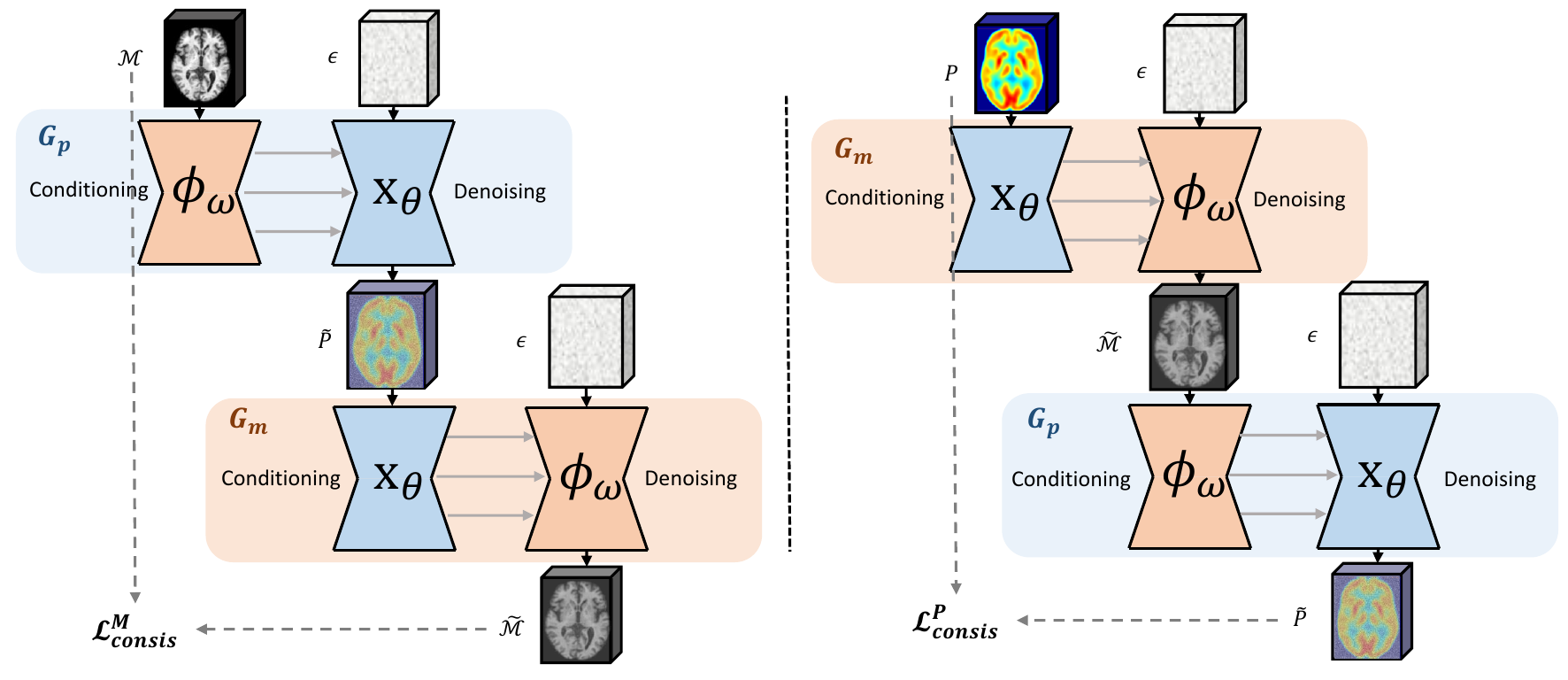}
    \caption{Cycle exchange consistency (CycleEx) strategy of \modelName. The two translation mappings $\boldsymbol{G}_p: \mathcal{M} \rightarrow \mathcal{P}$ and $\boldsymbol{G}_m: \mathcal{P} \rightarrow \mathcal{M}$ maintain cycle consistency. In addition, their network architectures are mirrored: both contain the same conditioner arm $\phi_\omega$ and denoiser arm $\mathbf{x}_\theta$, but with an exchanged position. This strategy ensures information sharing between the two arms.}
    \label{fig:cycle}
\end{figure}

The \modelName~framework further goes through a cycle exchange consistency (CycleEx) training strategy as shown in Fig.~\ref{fig:cycle}. The CycleEx stems from the cycle consistency loss proposed by Zhou \textit{et al.}~\citep{CycleGAN}, where it is claimed that the learned translation mapping functions should be cycle-consistent: for each image $\mathcal{M}^i$ from the MRI domain, given two mappings $\boldsymbol{G}_p: \mathcal{M} \rightarrow \mathcal{P}$ and $\boldsymbol{G}_m: \mathcal{P} \rightarrow \mathcal{M}$, the image translation cycle should be able to bring $\mathcal{M}^i$ back to its original input, i.e. $\mathcal{M}^i \rightarrow \boldsymbol{G}_p(\mathcal{M}^i) \rightarrow \boldsymbol{G}_m(\boldsymbol{G}_p(\mathcal{M}^i)) \approx \mathcal{M}^i$, as forward cycle consistency. Similarly, for each image $\mathcal{P}^i$ from the PET domain, $\boldsymbol{G}_p$ and $\boldsymbol{G}_m$ should also satisfy backward cycle consistency: $\mathcal{P}^i \rightarrow \boldsymbol{G}_m(\mathcal{P}^i) \rightarrow \boldsymbol{G}_p(\boldsymbol{G}_m(\mathcal{P}^i)) \approx \mathcal{P}^i$. In our CycleEx, the mapping $\boldsymbol{G}_m$ shares the same network architecture as $\boldsymbol{G}_p$, differing only in the swapped conditioner and denoiser arms. Therefore, during the forward cycle $\mathcal{M} \rightarrow \mathcal{P} \rightarrow \mathcal{M}$, the conditioner arm used to map input MRI to task-specific representations $\boldsymbol{h}_m$ during mapping $\boldsymbol{G}_p$, will be reused as the denoiser arm to synthesize MRI during mapping $\boldsymbol{G}_m$; the denoiser arm used to generate clean PET during mapping $\boldsymbol{G}_p$ will be reused to map input PET to task-specific representations $\boldsymbol{h}_p$ in mapping $\boldsymbol{G}_m$. The backward cycle $\mathcal{P} \rightarrow \mathcal{M} \rightarrow \mathcal{P}$ acts similarly. Thanks to the symmetric nature of the conditioner and denoiser arms in \modelName, this exchange can be achieved seamlessly. This design ensures that a single U-Net is dedicated to processing one specific modality consistently. 

The CycleEx procedure introduces three more conditional diffusion processes, without adding additional learnable model parameters. Each cycle introduces an additional cycle-consistency loss, bringing the training objective:
\begin{align}
    \mathcal{L}_{cycle}(\omega, \theta) &= \mathcal{L}_{consist}^\mathcal{M} + \mathcal{L}_{consist}^\mathcal{P}   \nonumber \\   
    &=\mathbb{E}_{\mathcal{M}}dist (\boldsymbol{G}_m (\boldsymbol{G}_p(\mathcal{M})), \mathcal{M}) + \mathbb{E}_{\mathcal{P}}dist(\boldsymbol{G}_p(\boldsymbol{G}_m(\mathcal{P})), \mathcal{P}), \tag{\textcolor{modicolor}{11}}
\end{align}
where $dist(\cdot)$ can be $L_1$ or $L_2$ norm. This setup enforces the information sharing between the two arms, bringing additional supervision and regularization to the image translation.


Finally, the combined training objective of \modelName~is:
\begin{equation}
    \mathcal{L} = \lambda_{task} * \mathcal{L}_{task} + \lambda_{diff} * \mathcal{L}_{diff} + \lambda_{cycle} * \mathcal{L}_{cycle}, \tag{\textcolor{modicolor}{12}}
\end{equation}
where $\lambda_{task}$, $\lambda_{diff}$, and $\lambda_{cycle}$ are constant multiplication factors that determine the relative importance of different losses. This joint training enables \modelName~to learn task-specific representations that preserve both structural and pathological details from the provided input modality, facilitating the generation of unseen modality with enhanced pathology-awareness. Moreover, CycleEx elevates translation quality by promoting the extraction of more informative features through effective dual-arm information exchange. The efficacy of individual parts will be shown in \Cref{sec:ablation}.

\subsubsection{Volumetric Generation}

A 2.5D generation strategy is introduced in \modelName~for memory-efficient 3D medical volumetric synthesis. Although a full 3D network offers comprehensive spatial understanding with inherent learning of inter-slice dependency, it has limitations due to its high computational demands. Moreover, the limited availability of paired, multi-modal medical data hinders the proper training of a large-volume 3D network. 
To strike a balance between training efficiency and consistent 3D PET generation, we adopt 2D convolutional layers in the neural network providing 2D slices as input, but cater the input channel with $N$ consecutive neighboring slices of the input slice along the same direction. After training, the network produces the target slice and its $N$ neighbors for each designated slice position. All neighboring slices are assigned a weight based on their distance from the target position, with central slices weighted highest and decreasing linearly for farther slices.
After summing these weighted slices, we average the overlapping accumulations for each slice position.
\textcolor{modicolor}{As all PET scans were resampled during preprocessing to a standardized template with isotropic voxel dimensions along all axes, the inter-slice spacing is uniform in physical space, ensuring that the output scans accurately reflect true anatomical separation, }
resulting in a balanced and consistent final 3D scan.
This strategy enables the model to process and recognize inter-slice relationships, efficiently mitigating the slice inconsistencies from the 2D network.




\section{Experimental Setup}
\label{sec:experiments}


\subsection{Datasets and Preprocessing} 
\begin{figure}[t]
    \centering
    \includegraphics[width=0.75\linewidth]{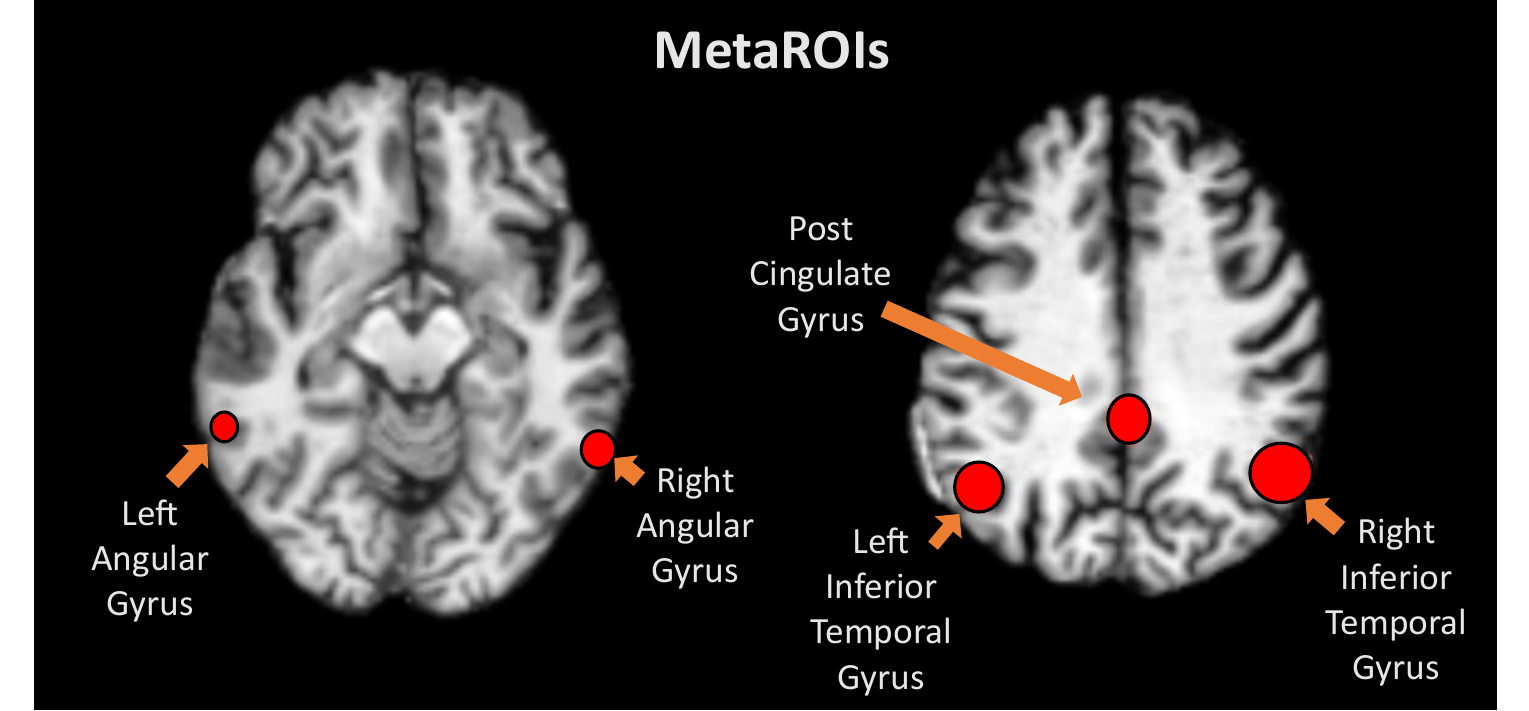}
    \caption{MetaROIs illustration on a brain MRI.}
    \label{fig:MetaROIs}
\end{figure}

We use paired T1-weighted MRI and PET scans from two different datasets for evaluation: \\
\indent 1) 1,248 subjects from the Alzheimer’s disease neuroimaging initiative (ADNI) database~\citep{adni}~(adni.loni.usc.edu), including cognitively normal (CN, n = 379) subjects, subjects with mild cognitive impairment (MCI, n = 611), and Alzheimer’s disease (AD, n = 257). \\
\indent 2) 253 subjects from a well-characterized, single-site in-house clinical dataset from TUM Klinikum, Munich, Germany. It includes 143 CN and 110 AD samples. \\
\indent All scans are additionally processed using SPM12\footnote{https://www.fil.ion.ucl.ac.uk/spm/software/spm12}. 
PET scans are normalized and registered to the MNI152 template with $1.5^3~mm^3$ voxel size. We further perform skull-stripping on all PET scans using Synthstrip~\citep{synthstrip} and MRI scans using Freesurfer~\citep{fischlFreeSurfer2012}. All data are min–max rescaled to the image intensity values between 0 and 1. All MRI scans are registered to the corresponding PET scans.
\textcolor{modicolor}{The rigid registration between MRI and PET is performed individually within each subject to align the modalities, without registration parameters sharing between different samples, avoiding leakage of spatial priors}.
The final image size for both modalities is $113 \times 137 \times 113$. To eliminate most blank backgrounds, we further center-crop all scans to $96 \times 112 \times 96$. We choose the axial direction of the brain scan as input. The MetaROIs mask is also registered to the MNI152 template with $1.5 ^3~mm^3$ voxel size. 
We demonstrate the five hypometabolic sensitive regions from the MetaROIs on a brain MRI in Fig.~\ref{fig:MetaROIs}.
For the ADNI data, the clinical data $\boldsymbol{c}$ includes six variables: age, gender, education, cognitive examination scores MMSE and ADAS-Cog-13, and genetic risk factor ApoE4. 
For the in-house clinical data, only age and gender are available.
Each group of variables is further standardized to a mean of 0 and a standard deviation of 1, prior to integration into the network.
We split the data into train, validation, and test sets,
\textcolor{modicolor}{including only one single scan per subject, acquired at the baseline visit, ensuring a strict subject-level separation between training, validation, and test sets.}
We also ensure that diagnosis, age, and gender are balanced across sets. We adapt the data splitting method from ClinicaDL~\citep{clinicadl}, where the balance of a split is first assessed by computing the propensity score --- the probability of a sample belonging to the training set --- using a logistic regression model consisting of the known confounders.
The percentiles of the propensity score distributions across the training, validation, and test sets are then compared, with the maximum deviation across all percentiles serving as a measure of imbalance~\citep{ho2007matching}. This procedure is repeated for 1,000 randomly selected partitions, and the partition with the lowest imbalance is ultimately selected. 
As a result, for ADNI data, we have 999 paired scans in the training set, 126 in the validation set, and 123 in the test set; for the in-house data, we use 165 for training, 39 for validation, and 49 for testing. 
\textcolor{modicolor2}{To prevent any bias due to the single split on the limited in-house data, we conducted and reported the results of full 5-fold cross-validation. However, for the larger ADNI dataset, we reported the results using one balanced split generated with the random seed of 666, to limit computational overhead.}


\subsection{Models and Hyperparameters} We adopt the ADM architecture~\citep{DMbeatsGAN} for both the conditioner and the denoiser arms in \modelName. 
From highest to lowest resolution, the UNet stages use $[C, 2C, 3C, 4C]$ channels, respectively. We set $C = 64$, depth $= 4$, and timestep $T = 1000$ with a cosine noise scheduler. We use global attention at downsampling factors $[16, 8, 4]$, which corresponds to attention on spatial resolutions $24 \times 28$, $12 \times 14$, and $6 \times 7$, given a $96 \times 112$ axial-direction input slice. 
We set the number of input neighboring slices $N = 15$, $\lambda_{task} = 0.1$, $\lambda_{diff} = 1$, $\lambda_{cycle} = 1$ for the training objective after an exhaustive search. $L_1$ loss is used for both the denoising and the predefined task loss. The model results in $89M$ parameters.

We use the AdamW optimizer with a learning rate of $5 \times 10^{-4}$, a weight decay of $10^{-6}$, $(\beta_1, \beta_2) = (0.9, 0.999)$, a batch size of $6$, and an exponential moving average (EMA) over model parameters with a rate of $0.999$. We adopt the DDIM~\citep{ddim} strategy with 100-step sampling. We train the network with an NVIDIA A100 GPU for 72K iterations, with a duration of $96$ hours. 
For the classification task, we use a 3D ResNet for all modalities input. The classifier is trained using the AdamW optimizer with a learning rate of $0.005$, a weight decay of $10^{-6}$, and a batch size of 32 for 5K iterations.

\subsection{Baselines} Our baseline methods include Pix2Pix~\citep{pix2pix}, CycleGAN~\citep{CycleGAN}, \textcolor{modicolor}{RegGAN~\citep{kong2021breaking}}, ResVit~\citep{resvit}, BBDM~\citep{bbdm}, and BBDM-LDM~\citep{bbdm}. 
ResVit~\citep{resvit} is a GAN-based method that integrates ResNet and ViT as backbones, designed for medical image translation. 
Unfortunately, other GAN-based MRI to PET translation approaches do not provide open-source code~\citep{ganbert,gandalf,sketcher-refine}. 
To ensure the inclusion of representative GAN-based methods in our comparison, we implemented the widely-used image translation techniques Pix2Pix~\citep{pix2pix} and CycleGAN~\citep{CycleGAN}\textcolor{modicolor}{, as well as RegGAN~\citep{kong2021breaking}, a medical image translation approach designed primarily for multi-contrast MRI translation}.
While none of the state-of-the-art DM-based translation methods has been proposed for MRI to PET translation, BBDM~\citep{bbdm}, a method modeling image translation as a stochastic Brownian Bridge process, stands out for its exceptional replicability and performance. Thus we select it for adaptation and comparison.
We also include its variation BBDM-LDM, which is based on the latent diffusion models (LDM)~\citep{stable_diffusion}. We use the same training and evaluation data in all baseline methods as PASTA. 

\subsection{Evaluation Setup} We conduct a comprehensive evaluation of all methods both qualitatively and quantitatively. 
The following metrics between the GT and synthesized PET are computed:
mean absolute error (MAE$\downarrow$), mean squared error (MSE$\downarrow$), peak signal-to-noise ratio (PSNR$\uparrow$), and structure similarity index (SSIM$\uparrow$). To further validate the preservation of pathology on the synthesized PET, we implement a downstream AD classification task with 5-fold cross-validation. We use balanced accuracy (BACC), F1-Score, and area under the ROC curve (AUC) to evaluate the classification results. For qualitative assessment, we present our generative results, additional results on 3D-SSP maps, as well as the clinical evaluation from our collaborated physicians. We further investigate the impact of individual clinical data and fairness evaluation.
\textcolor{modicolor}{The evaluation of the computational costs of all methods can be found in~\ref{sec:computational_costs}.}


\section{Results and Discussion}
\label{sec:results}

\subsection{Qualitative Results and Clinical Evaluation}

\begin{figure}[t]
    \centering
    \includegraphics[width=\linewidth]{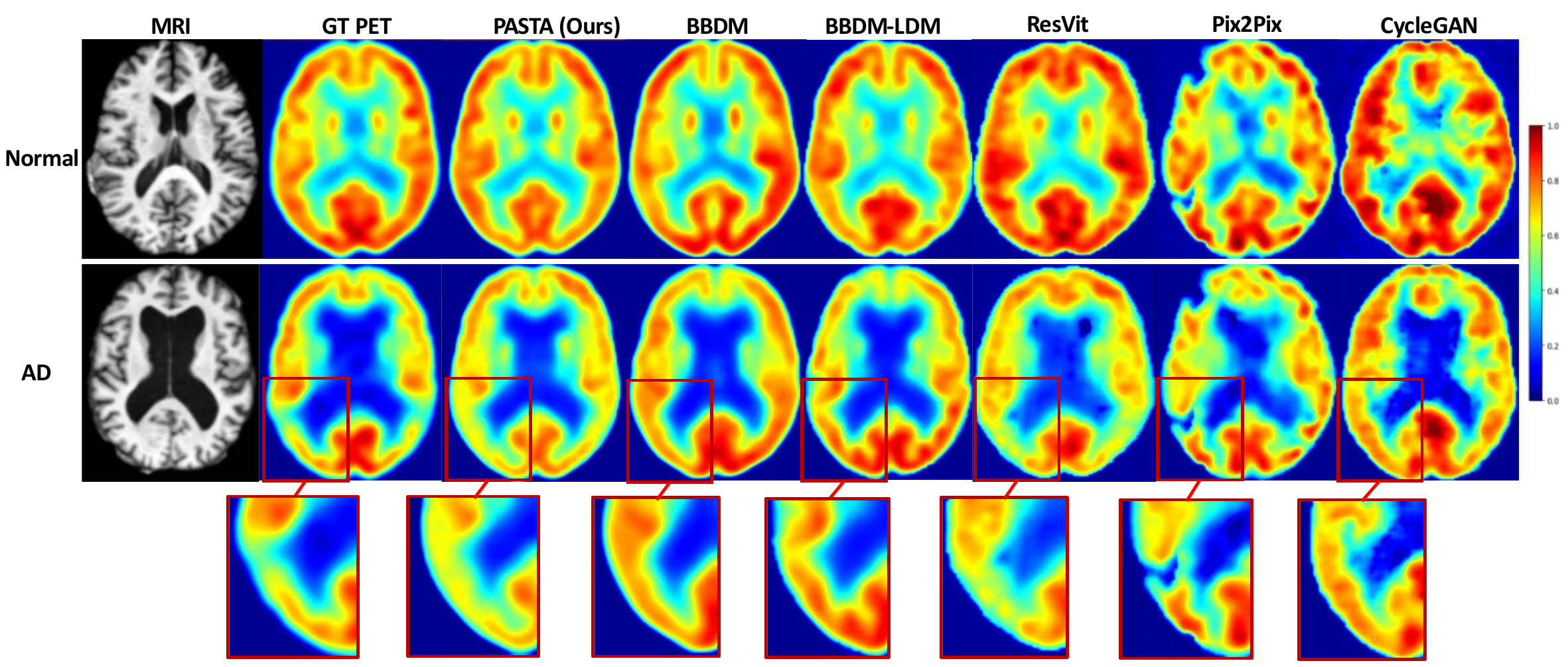}
    \caption{Qualitative comparison of cross-modality synthesis methods. The first row shows images from a normal control subject with no obvious pathology, and the second row shows an AD patient with obvious hypometabolism in the temporoparietal lobe (bottom left and right parts). The left temporoparietal lobe is magnified in the third row. For the normal subject, most baselines recover the structure and metabolic information well; for the AD subject, \modelName~demonstrates superior generation fidelity and pathology preservation.}
    \label{fig:bl_comp}
\end{figure}

Fig.~\ref{fig:bl_comp} presents qualitative comparisons of MRI to PET translation results for \modelName~and state-of-the-art baseline methods on the ADNI dataset.
These visual results, analyzed in cooperation with our clinical experts, clearly show that \modelName~produces PET scans with higher fidelity and closer resemblance to the ground truth (GT) PET than other methods. For methods Pix2pix and CycleGAN, the generated scans deviate significantly from the GT PET scans, with noticeable artifacts and inconsistencies. For AD patients, the generated PET scans from \modelName~accurately capture the reduced metabolism in the temporoparietal lobe, a key region highly associated with AD, as observed in the GT PET scan. Another DM-based model, BBDM, effectively preserves the structural details but struggles to correctly translate pathological information. Its LDM-based variation, BBDM-LDM, also falls short in maintaining pathology. ResVit, a model developed specifically for medical translation, improves pathology awareness but sacrifices accuracy in anatomical structures. Overall, \modelName~demonstrates superior performance, achieving a remarkable balance between preserving both structural and pathological details consistent with the GT scans.

To validate PASTA's clinical efficacy, we sought feedback from our clinical collaborators regarding the fidelity and pathological accuracy of the generated PET scans. According to their evaluations, our generated PET scans are realistic and comparable to the real PET. 
While the generated scans appear generally smoother, this is not considered a concern in clinical practice, as nuclear physicians commonly apply filters to PET images, and AD diagnosis does not rely on high-resolution edge details~\citep{pet-hypometa}. 
The synthesized PET scans for AD patients show pathological patterns consistent with actual data, albeit less pronounced. This is expected, as PET synthesis relies on structural MRI, a modality less sensitive to functional neurodegenerative abnormalities. Nonetheless, the synthesized PETs still exhibit higher pathological sensitivity for AD diagnosis compared to their corresponding MRI inputs.

\subsection{Quantitative Results}

We report the quantitative metrics measuring the absolute errors and structural similarities between GT and synthesized PET for different methods in Table~\ref{tab:bl-comp-adni} \textcolor{modicolor}{and Table~\ref{tab:bl-comp-inhouse-2}}.
On the ADNI dataset, consistent with the qualitative results, \modelName~generates PET scans with the highest quality, obtaining the lowest MAE (\textcolor{modicolor2}{0.0345 ± 0.0051}), MSE (\textcolor{modicolor2}{0.0043 ± 0.0010}), and highest PSNR (\textcolor{modicolor2}{24.59 ± 0.88}), SSIM (\textcolor{modicolor2}{86.29 ± 1.19\%})\textcolor{modicolor2}{, represented as the mean ± standard deviation of the quantitative metrics across the whole test samples}.  
The DM-based method BBDM consistently has the second-best results, with its LDM-based variation BBDM-LDM similar performance. However, the other GAN-based baselines, especially CycleGAN, cannot reach on-par performance. The results on the in-house dataset are generally slightly lower compared to the ADNI dataset, likely due to the limited amount of training data available. 
\textcolor{modicolor}{To prevent any bias due to the single split on the limited in-house data, we further conducted a 5-fold cross-validation on different methods. Results shown in Table~\ref{tab:bl-comp-inhouse} deliver similar overall trend as observed in the ADNI dataset, with PASTA obtaining the lowest MAE of 0.0430 ± 0.0008, MSE of 0.0061 ± 0.0002, and the highest PSNR of 23.20 ± 0.06, SSIM of 85.4 ± 0.4\%. Compared to the other baseline methods, PASTA also achieved the smallest variation among different folds across four metrics, indicating its stable and robust performance.}
\textcolor{modicolor2}{Pairwise statistical comparisons were performed between PASTA and all baseline methods using the two-sided Wilcoxon signed-rank test on all quantitative metrics. Across both datasets, the results consistently yielded a $p$-value $< 0.0001$, demonstrating that the performance improvement gained by PASTA is statistically significant over the baseline methods.}
These results confirm the potential of diffusion models in medical image translation.

\begin{table}[t]
\small
\centering
{\color{modicolor2}
{\setlength{\tabcolsep}{0.4em}
\begin{tabular}{llcccc}\toprule
Data & Method & MAE\scriptsize{($\times 10^{-2}$)}$\downarrow$  & MSE\scriptsize{($\times 10^{-2}$)}$\downarrow$ & PSNR$\uparrow$ & SSIM(\%)$\uparrow$ \\\midrule
\multirow{7}{*}{\rotatebox{90}{ADNI}} 
& CycleGAN \citep{CycleGAN} 
        & 9.83 \interval{0.92} 
        & 2.61 \interval{0.42} 
        & 16.38 \interval{0.66} 
        & 47.48 \interval{2.36}  \\
        
    & Reg-GAN \citep{kong2021breaking} 
        & 8.66 \interval{1.03}
        & 2.47 \interval{0.48}
        & 16.66 \interval{0.83}
        & 55.70 \interval{1.31} \\
        
    & Pix2Pix \citep{pix2pix}    
        & 5.56 \interval{0.43} 
        & 1.15 \interval{0.14} 
        & 19.90 \interval{0.49} 
        & 73.23 \interval{1.42} \\
        
    & ResVit \citep{resvit} 
        & 6.72 \interval{0.72} 
        & 1.74 \interval{0.24} 
        & 19.40 \interval{0.59} 
        & 69.83 \interval{2.08} \\
        
    & BBDM-LDM \citep{bbdm}  
        & 3.96 \interval{0.69} 
        & \underline{0.54 \interval{0.15}} 
        & \underline{23.75 \interval{1.01}} 
        & 84.25 \interval{1.30} \\
        
    & BBDM \citep{bbdm}  
        & \underline{3.88 \interval{0.47}} 
        & 0.56 \interval{0.11} 
        & 23.37 \interval{0.76} 
        & \underline{84.55 \interval{1.36}} \\\cmidrule{2-6}
        
    & PASTA (Ours)
        & \textbf{3.45 \interval{0.51}} 
        & \textbf{0.43 \interval{0.10}} 
        & \textbf{24.59 \interval{0.88}} 
        & \textbf{86.29 \interval{1.19}} \\
    \bottomrule
    \end{tabular}}}
    
    \caption{\textcolor{modicolor2}{Quantitative comparison between the baselines on the ADNI dataset. We report the mean $\pm$ standard deviation across the test set.}}
 \label{tab:bl-comp-adni}
\end{table}

\begin{table}[t]
\small
\centering
{\color{modicolor2}
{\setlength{\tabcolsep}{0.4em}
\begin{tabular}{llcccc}\toprule
Data & Method & MAE\scriptsize{($\times 10^{-2}$)}$\downarrow$  & MSE\scriptsize{($\times 10^{-2}$)}$\downarrow$ & PSNR$\uparrow$ & SSIM(\%)$\uparrow$ \\\midrule
\multirow{7}{*}{\rotatebox{90}{In-house}} 
& CycleGAN \citep{CycleGAN} 
        & 15.10 \interval{1.40}
        & 4.76 \interval{1.06}
        & 15.29 \interval{0.68}
        & 29.30 \interval{2.50} \\
        
    & Reg-GAN \citep{kong2021breaking}
        & 10.83 \interval{1.52}
        & 2.47 \interval{0.52}
        & 16.17 \interval{0.91}
        & 42.02 \interval{2.25} \\
        
    & Pix2Pix \citep{pix2pix}
        & 9.27 \interval{1.62}
        & 2.32 \interval{0.93}
        & 18.50 \interval{0.95}
        & 58.37 \interval{5.68} \\
        
    & ResVit \citep{resvit}
        & 6.05 \interval{1.48}
        & 1.31 \interval{0.93}
        & 19.18 \interval{1.37}
        & 65.86 \interval{5.17} \\
        
    & BBDM-LDM \citep{bbdm}
        & \underline{4.66 \interval{2.42}}
        & 0.73 \interval{0.50}
        & 22.71 \interval{4.36}
        & \underline{79.88 \interval{7.25}} \\
        
    & BBDM \citep{bbdm}
        & 4.85 \interval{1.69}
        & \underline{0.61 \interval{0.39}}
        & \underline{23.03 \interval{3.11}}
        & 52.07 \interval{16.9} \\
    \cmidrule{2-6}
        
    & PASTA (Ours)
        & \textbf{4.22 \interval{0.74}}
        & \textbf{0.60 \interval{0.19}}
        & \textbf{24.42 \interval{1.21}}
        & \textbf{85.90 \interval{2.02}} \\
\bottomrule
\end{tabular}}}

\caption{\textcolor{modicolor2}{Quantitative comparison between the baselines on the in-house dataset. We report the mean $\pm$ standard deviation across one balanced test split.}}
 \label{tab:bl-comp-inhouse-2}
\end{table}

\begin{table}[t]
\small
\centering
{\color{modicolor}
{\setlength{\tabcolsep}{0.4em}
\begin{tabular}{llcccc}\toprule
Data & Method & MAE\scriptsize{($\times 10^{-2}$)}$\downarrow$  & MSE\scriptsize{($\times 10^{-2}$)}$\downarrow$ & PSNR$\uparrow$ & SSIM(\%)$\uparrow$ \\\midrule
\multirow{7}{*}{\rotatebox{90}{In-house}} 
& CycleGAN \citep{CycleGAN} & 11.7 \interval{1.91} & 2.88 \interval{0.95} & 16.09 \interval{1.14} & 31.8 \interval{6.6} \\
& RegGAN \citep{kong2021breaking} & 10.8 \interval{1.41} & 2.53 \interval{0.65} & 16.91 \interval{0.82} & 41.5 \interval{2.6} \\
& Pix2Pix \citep{pix2pix} & 8.75 \interval{0.40} & 2.03 \interval{0.11} & 18.73 \interval{0.19} & 58.3 \interval{2.4} \\
& ResVit \citep{resvit} & 7.09 \interval{0.74} & 1.87 \interval{0.34} & 19.93 \interval{0.48} & 68.3 \interval{2.5} \\
& BBDM-LDM \citep{bbdm} & \underline{4.48} \interval{0.17} & \underline{0.66} \interval{0.05} & \underline{23.04} \interval{0.37} & \underline{78.9} \interval{0.7} \\
& BBDM \citep{bbdm} & 4.70 \interval{0.62} & 0.69 \interval{0.05} & 22.88 \interval{0.67} & 71.0 \interval{12.0} \\\cmidrule{2-6}
& PASTA (Ours) & \textbf{4.30} \interval{0.08} & \textbf{0.61} \interval{0.02} & \textbf{23.20} \interval{0.06} & \textbf{85.4} \interval{0.4} \\
\bottomrule
\end{tabular}}}

\caption{\textcolor{modicolor}{Quantitative comparison between the baselines on the in-house dataset with 5-fold cross-validation.
\textcolor{modicolor2}{We report the mean $\pm$ standard deviation across the five test splits.}
}}
 \label{tab:bl-comp-inhouse}
\end{table}

\begin{table}[b]
 \centering
 \small
 
 {\setlength{\tabcolsep}{0.7em}
 \begin{tabular}{lccc}\toprule
    Input & BACC $\uparrow$ & F1-Score $\uparrow$ & AUC $\uparrow$  \\\midrule
    MRI   & 79.23 \interval{4.30} & 74.97 \interval{5.95} & 85.88 \interval{3.79} \\
    MRI+$\mathbf{c}$ & 81.51 \interval{3.16} & 77.83 \interval{3.83} & \underline{89.19} \interval{1.41} \\
    GT PET & {\textbf{87.02}} \interval{2.35} & \textbf{80.77} \interval{2.62} & 89.04 \interval{1.88} \\
    Syn PET (ResVit) & 78.54 \interval{4.15} & 74.41 \interval{5.28} & 81.80 \interval{4.95} \\
    Syn PET (BBDM) & 72.93 \interval{8.48} & 70.76 \interval{6.02} & 83.29 \interval{5.13} \\
    Syn PET (\modelName) & \underline{83.41} \interval{2.67} & \underline{79.98} \interval{3.51} & \textbf{91.63} \interval{2.21} \\
    \bottomrule
 \end{tabular}}
 \caption{Classification results for Alzheimer's disease classification with different input modalities.}
 \label{tab:classification}
\end{table}

\subsection{Classification Results for AD Diagnosis}
To assess the value of generated 3D PET scans in AD diagnosis, we train AD classifiers on MRI, MRI with clinical data (MRI+$\mathbf{c}$), GT PET, and synthesized (Syn) PET scans, respectively. 
We use ResVit, BBDM and \modelName~for the PET generation. 
A 3D ResNet is employed as the classifier to evaluate all modalities input. 
To mitigate potential domain shift issues and ensure fair comparison, we train and test classifiers on images from the same source. We also use the same data splits for the classifier as for \modelName.
Table~\ref{tab:classification} reports the classification results on the ADNI dataset, and, as expected, GT PET has a higher performance than MRI across all metrics. 
The results for \modelName~also improve over MRI across all metrics with an increase of over 4\%. 
\textcolor{modicolor}{Statistical significance testing further supports this improvement. The McNemar's test on BACC yielded a $p$-value of 0.022, with a bootstrapped 95\% confidence interval (CI) for the difference ranging from 1.61\% to 10.6\%. Similarly, the DeLong test on AUC scores produced a $p$-value of 0.00016, with a bootstrapped 95\% CI for the difference between 3.20\% and 10.7\%.}
The inclusion of clinical data only improved MRI's BACC by 2\%, suggesting that \modelName's enhanced pathology awareness is likely attributed to its more effective learning of the interaction between MRI and clinical data. 
While the results for \modelName~in BACC fall between those of MRI and GT PET, it almost matches GT PET in F1-Score and achieves the highest AUC. 
In contrast, results for BBDM are worse than those of MRI, confirming its limitations in pathology transfer, consistent with the qualitative evaluation in Fig.~\ref{fig:bl_comp}. The ResVit results show improved accuracy compared to BBDM, attributed to its better pathology awareness demonstrated in the qualitative results, yet are still lower than for MRI.
These results underscore the high potential of \modelName~for AD diagnosis and highlight the necessity of pathology-aware transfer. 


\subsection{\textcolor{modicolor2}{Pathology-Localized Evaluation}}

\textcolor{modicolor2}{
To assess pathology preservation beyond global image fidelity, we evaluated all methods using pathology-localized metrics computed within AD-related MetaROIs~\citep{metaROIs} regions, namely MAE$_{ROI}$, MSE$_{ROI}$, PSNR$_{ROI}$, and SSIM$_{ROI}$. As shown in \Cref{tab:roi-comp-adni}, PASTA consistently outperforms all baseline methods across all ROI-based metrics on the ADNI dataset, achieving the lowest MAE$_{ROI}$ of $9.07 \pm 2.21 \times 10^{-4}$, the highest PSNR$_{ROI}$ of $38.29 \pm 1.65$ and SSIM$_{ROI}$ of $99.74 \pm 0.07$\%. This demonstrates its superior generation fidelity specifically within disease-relevant regions.
Across all metrics, PASTA and its variants significantly outperform competing baseline methods ($p$-values $< 0.0001$, two-sided Wilcoxon signed-rank test).
}

\begin{table}[t]
\small
\centering
{\color{modicolor2}
{\setlength{\tabcolsep}{0.2em}
\begin{tabular}{lcccc}\toprule
Method 
& MAE$_{ROI}$\scriptsize{($\times 10^{-4}$)}$\downarrow$ 
& MSE$_{ROI}$\scriptsize{($\times 10^{-4}$)}$\downarrow$ 
& PSNR$_{ROI}$$\uparrow$ 
& SSIM$_{ROI}$\scriptsize{(\%)}$\uparrow$ \\
\midrule
CycleGAN \citep{CycleGAN}
    & 18.54 \interval{4.94}
    & 4.51 \interval{2.23}
    & 32.73 \interval{2.13}
    & 99.41 \interval{0.16} \\

Reg-GAN \citep{kong2021breaking}
    & 21.06 \interval{4.03}
    & 6.69 \interval{1.86}
    & 30.88 \interval{0.93}
    & 99.09 \interval{0.10} \\

Pix2Pix \citep{pix2pix}
    & 12.59 \interval{2.14}
    & 2.29 \interval{0.64}
    & 35.46 \interval{1.21}
    & 99.52 \interval{0.06} \\

ResVit \citep{resvit}
    & 13.12 \interval{2.20}
    & 2.86 \interval{0.69}
    & 34.49 \interval{1.34}
    & 99.42 \interval{0.11} \\

BBDM-LDM \citep{bbdm}
    & 10.79 \interval{2.36}
    & 1.63 \interval{0.63}
    & 37.03 \interval{1.69}
    & 99.67 \interval{0.08} \\

BBDM \citep{bbdm}
    & 10.54 \interval{2.24}
    & 1.66 \interval{0.61}
    & 36.96 \interval{1.70}
    & 99.64 \interval{0.09} \\
\cmidrule{1-5}

PASTA (Ours)
    & \textbf{9.07 \interval{2.21}}
    & \textbf{1.24 \interval{0.44}}
    & \textbf{38.29 \interval{1.65}}
    & \textbf{99.74 \interval{0.07}} \\


\bottomrule
\end{tabular}}}

\caption{\textcolor{modicolor2}{Pathology-localized quantitative comparison on the ADNI dataset using ROI-based metrics computed within AD-related MetaROI regions. We report the mean $\pm$ standard deviation across the test set.}}
\label{tab:roi-comp-adni}
\end{table}

\subsection{Influence of the Clinical Data}
\label{sec:clinicaldata}

To investigate the impact and sensitivity of individual clinical variables from the ADNI dataset on the generation process, we conducted an additional experiment during inference. For each of the six clinical variables: Age, Gender, Education level (Edu), MMSE, ADAS-Cog-13 (ADAS), and ApoE4, we retained its original value while neutralizing the other five by setting them to their respective mean values derived from the dataset. The outcomes were then compared against the original configuration, where all variables were preserved (denoted as ``All''). To introduce a stronger perturbation, CN subjects are assigned the mean of the AD group and vice versa. 
\Cref{tab:clinicaldatainf} reports the mean absolute error (MAE, $\times 10^{-2}$) for the generation across the test set. To quantitatively assess their pathology influence, we also present the MAE within the AD-related MetaROIs~\citep{metaROIs} regions ($\text{MAE}_{ROI}$, $\times 10^{-4}$). 
ADAS emerges as the most influential variable, showing the smallest deviation from the baseline when preserved alone for both CN and AD subjects, followed by MMSE. While other variables, such as Age, Education, and ApoE4, are comparatively less impactful.

\begin{table}[h]
    \centering
    \small
    
    {\setlength{\tabcolsep}{0.6em}
    \begin{tabular}{llrrrrrrr}\toprule
       &   & Age & Gender & Edu & MMSE & ADAS & ApoE4 & All \\\midrule
    \multirow{2}{*}{CN} & \scriptsize{MAE}$\downarrow$ & 3.34 & 3.32 & 3.34 & 3.30 & 3.30 & 3.34 & 3.26 \\
     & \scriptsize{$\text{MAE}_{ROI}$}$\downarrow$ & 9.82 & 9.61 & 9.68 & 9.10 & 8.46 & 9.58 & 8.42 \\
     \midrule
     \multirow{2}{*}{AD} & \scriptsize{MAE}$\downarrow$ & 3.57 & 3.57 & 3.57 & 3.55 & 3.57 & 3.57 & 3.57 \\
     & \scriptsize{$\text{MAE}_{ROI}$}$\downarrow$ & 10.16 & 10.12 & 10.17 & 10.13 & 9.94 & 10.14 & 9.84 \\
     \bottomrule
    \end{tabular}}
    \caption{Results of sensitivity analysis, keeping one clinical variable at a time while neutralizing the rest during inference on ADNI test set. The outcomes are compared against the original configuration, where all variables were preserved (All). (MAE: $\times 10^{-2}$, $\text{MAE}_{ROI}$: $\times 10^{-4}$)}
    \label{tab:clinicaldatainf}
\end{table}

\begin{figure}[t]
    \centering
    \includegraphics[width=\linewidth]{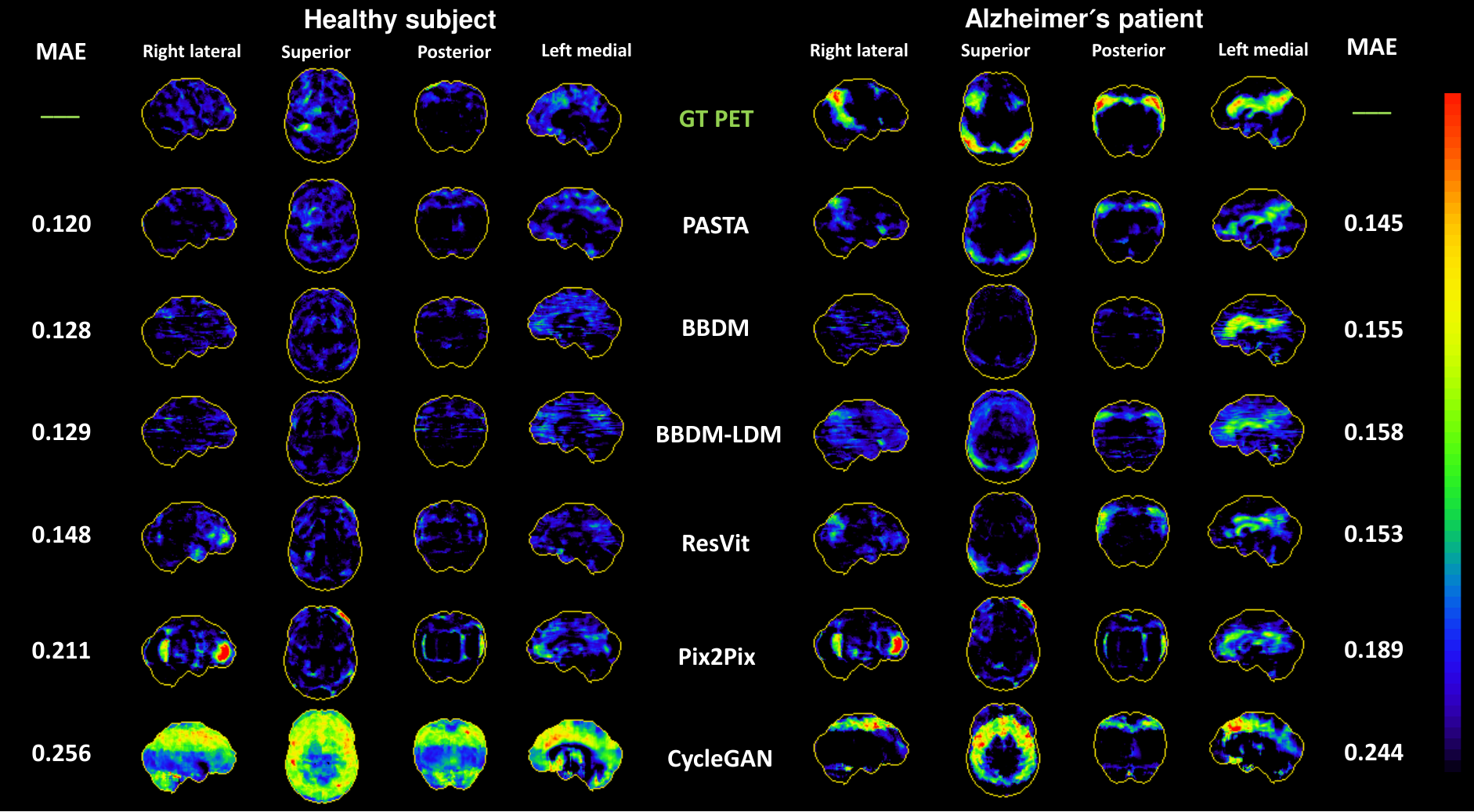}
    \caption{Z-score maps produced by Neurostat 3D-SSP for a healthy (left) and an Alzheimer's patient (right). We present global metabolic Z-score maps for the ground-truth (GT) PET and synthesized PET scans from each method in different brain cortical areas. The mean absolute errors (MAE) between the maps from synthesized PETs and the GT are displayed alongside.}
    \label{fig:ssp}
\end{figure}

\subsection{Neurostat 3D-SSP Maps for PET Scans}
\label{sec:neurossp}

Neurostat 3D-SSP (Neurological Statistical Image Analysis Software 3D Stereotactic Surface)~\citep{3dssp} is a statistical quantitative brain mapping tool designed to investigate brain disorders and assist clinical diagnosis using PET. Widely adopted in clinical settings, it has contributed to identifying functional abnormalities in various brain disorders and supports accurate diagnoses.
3D-SSP evaluates the statistical significance of differences in cortical metabolic activity between a patient and age-matched healthy controls.
The process involves spatially transforming trans-axial brain images to match a 3D reference brain from a stereotactic atlas, extracting peak cortical metabolic activity values, and projecting them onto a surface rendition of the brain. The resulting projections are statistically compared pixel-wise against a database of PET scans in age-matched controls, producing Z-score maps that highlight significant deviations~\citep{3d-ssp-explain}.
This tool provides a reliable quantitative method for evaluating the pathology consistency of synthesized PET scans. We present the generated Z-score maps by 3D-SSP for ground-truth (GT) PET, synthesized PET from \modelName~and baseline methods, as well as the mean absolute error between the GT Z-score maps and those from synthesized ones in Fig.~\ref{fig:ssp}. 
These maps are also used for clinical evaluation by our collaborating physicians.
For a healthy control subject, \modelName, BBDM, and BBDM-LDM produce metabolic patterns closely matching the GT PET, whereas ResVit, Pix2Pix, and CycleGAN introduce abnormalities absent in the actual data. For an Alzheimer’s patient, \modelName~accurately identifies pathological regions consistent with the GT PET, though with less pronounced patterns. Other DM-based models, BBDM and BBDM-LDM, fail to recover these pathological features. ResVit shows improved pathological recovery, yet its suboptimal performance in the healthy control subject raises concerns regarding its reliability. Pix2Pix and CycleGAN exhibit significant deviations from GT PET in both cases.
Overall, the 3D-SSP Z-score maps indicate that \modelName~achieves superior pathology awareness compared to other DM- and GAN-based image translation methods.

\subsection{Ablation Study}
\label{sec:ablation}

\subsubsection{Important Designs in the Proposed Framework}
\label{sec:ablation_important_designs}
We perform ablative experiments to verify the effectiveness of several important designs in our framework, including the CycleEx, integration of pathology priors ($\boldsymbol{\lambda}_{R}$), and clinical data ($\boldsymbol{c}$). Other ablation studies include: 1) a different way to integrate multi-modal conditions --- simply concatenating multi-scale features in the conditional module instead of using AdaGN for the multi-modal fusion, denoted as ConcatFeats; 2) a different predefined task in the conditioner arm, e.g., MRI reconstruction, denoted as M2M; 3) alternative integration positions for multi-modal conditions --- either exclusively within the conditioner arm (denoted as CondCond), or in both the conditioner and denoiser arms (denoted as CondBoth), rather than solely in the denoiser arm; 4) different loss weights ($\lambda_{task}$) for the predefined task;
\textcolor{modicolor2}{
5) making the constant weight factor $\lambda_R$ on the MetaROIs area in the pathology priors $\boldsymbol{\lambda}_{R}$ learnable, denoted as $\lambda_R$ learn;
}
\textcolor{modicolor}{
6) a new adaptive conditional module with slice-aware adaptive group normalization (denoted as SA-AdaGN), injecting slice position as priors into the generation process (see detailed description in~\ref{sec:sa-adagn});
7) adding an auxiliary classifier consistency loss (denoted as CCL) into the training objective to integrate discriminative signal from the disease label supervision (see detailed description in~\ref{sec:CCL}).
}
As shown in~\Cref{tab:ablation-designs}, CycleEx especially elevates the generation quality by a large margin. The choice of the predefined task in the conditioner arm and its loss weighting parameter also exhibit a relatively high impact.
\textcolor{modicolor}{SA-AdaGN and CCL deliver performance largely comparable to the baseline. 
\textcolor{modicolor2}{
Incorporating the learnable weight factor $\lambda_R$ within the MetaROIs prior on top of the CCL module offered a marginal improvement, leading to a slight improvement in the MAE and SSIM compared to adding CCL alone.}
Although they exhibit slightly higher MAE, the accompanying SSIM improvement suggests a potential advantage in preserving the structural fidelity of the translated PET.}

\begin{table}[t]
 \centering
 \small
 
 {\setlength{\tabcolsep}{0.6em}
 \begin{tabular}{lcccc}\toprule
    Ablation & MAE \scriptsize{($\times 10^{-2}$)}$\downarrow$  & MSE \scriptsize{($\times 10^{-2}$)}$\downarrow$ & PSNR$\uparrow$ & SSIM\scriptsize{(\%)}$\uparrow$
    \\\midrule
    w/o  CycleEx & 3.99 & 0.54 & 23.64 & 85.14 \\
    w/o $\boldsymbol{\lambda}_{R}$ & 3.67 & 0.48 & 24.19 & 85.95 \\
    w/o $\boldsymbol{c}$ & 3.57 & 0.46 & 24.39 &  86.12 \\
    PASTA (ConcatFeats) & 3.61 & 0.48 & 24.04 & 85.00 \\
    PASTA (M2M) & 3.78 & 0.50 & 23.78 & 84.77 \\
    PASTA (CondCond) & 3.55 & 0.45 & 24.32 & 86.11 \\
    PASTA (CondBoth) & 3.61 & 0.46 & 24.29 & 86.09 \\
    PASTA ($\lambda_{task}$ = 0.01) & 3.65 & 0.49 & 24.01 & 85.49 \\
    PASTA ($\lambda_{task}$ = 1) & 3.70 & 0.49 & 24.11 & 85.71 \\
    PASTA ($\lambda_{task}$ = 10) & 3.81 & 0.48 & 24.04 & 83.24 \\ 
    \textcolor{modicolor2}{PASTA ($\lambda_{R}$ learn)} & \textcolor{modicolor2}{4.27} & \textcolor{modicolor2}{0.62} & \textcolor{modicolor2}{23.08} & \textcolor{modicolor2}{86.13} \\
    \textcolor{modicolor}{PASTA (SA-AdaGN)} & \textcolor{modicolor}{3.81} & \textcolor{modicolor}{0.52} & \textcolor{modicolor}{23.77} & \textcolor{modicolor}{87.74} \\
    \textcolor{modicolor}{PASTA (w/ CCL)} & \textcolor{modicolor}{3.91} & \textcolor{modicolor}{0.53} & \textcolor{modicolor}{23.73} & \textcolor{modicolor}{88.01}
    \\
    \textcolor{modicolor2}{PASTA (w/ CCL, $\lambda_{R}$ learn)} & \textcolor{modicolor2}{3.89} & \textcolor{modicolor2}{0.53} & \textcolor{modicolor2}{22.95} & \textcolor{modicolor2}{\textbf{88.39}}
    \\
    \midrule
    \modelName  & \textbf{3.45} & \textbf{0.43} & \textbf{24.59} &  86.29
    \\\bottomrule
 \end{tabular}}
 \caption{Ablation studies on important designs in \modelName.}
 \label{tab:ablation-designs}
\end{table}

\begin{figure}[t!]
    \centering
    \includegraphics[width=\linewidth]{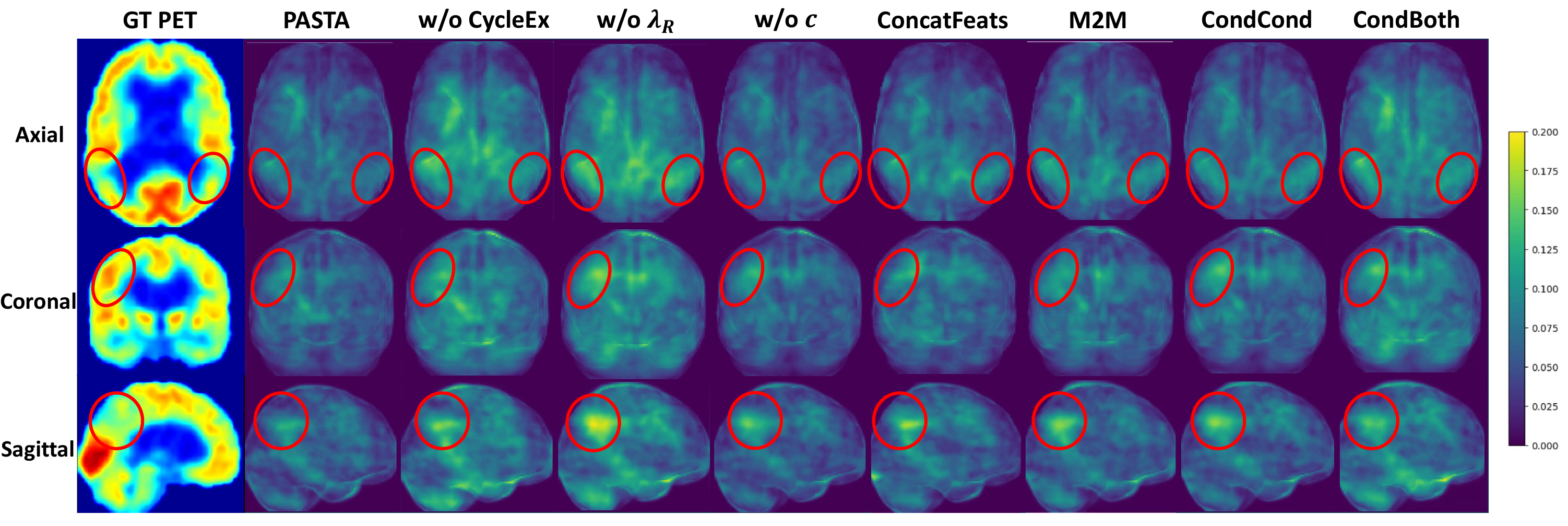}
    \caption{Error maps for ablation studies on important designs in \modelName, for an AD-diagnosed subject. We show its ground-truth (GT) PET and the averaged error maps of the scan along three directions respectively. Specifically, we circle the areas that are highly influenced by AD in red.}
    \label{fig:abla-errormap}
\end{figure}

In addition, we present error maps of generated PET scans from ablation studies on different design elements of \modelName~for an AD-diagnosed subject, compared to its ground-truth PET, in Fig.~\ref{fig:abla-errormap}. The results reveal that modifications to \modelName~consistently increase error magnitudes in regions with pathological features, particularly within the parietal lobe across the three anatomical planes, highlighted with red circles. Among the various factors, the pathology priors ($\boldsymbol{\lambda}_{R}$) and the CycleEx strategy exert a particularly pronounced impact on the quality of the generated outputs. This observation emphasizes the critical role of different components in \modelName~for enhancing the model's pathology awareness.


\subsubsection{Input Number of Neighboring Slices} 
\begin{figure}[t]
    \centering
    \includegraphics[width=1\linewidth]{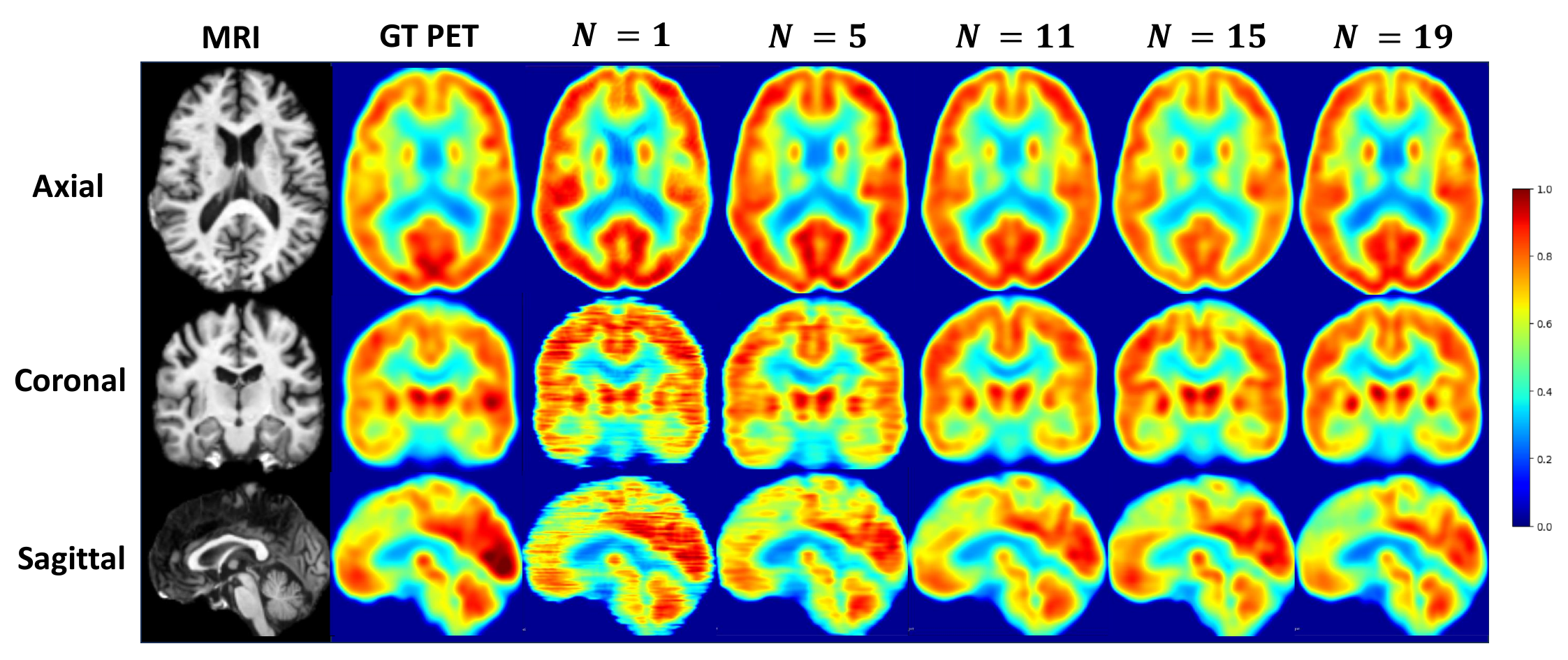}
    \caption{Ablation on input number $N$ of neighboring slices. We demonstrate the synthesized PET scans with various $N$s in three directions (axial, coronal, and sagittal), given that axial slices are used as input to the network.}
    \label{fig:ablation_N}
\end{figure}

The 2.5D strategy for the volumetric generation introduced by \modelName~caters to the input channel with $N$ consecutive neighboring slices of the input slice along the same direction. After training, the network also produces the target slice with its $N$ neighbors for each slice position, and the accumulated overlapping slices are linearly averaged for the final 3D scan. We evaluate the influence of the number $N$ of input neighboring slices on the quality of generated PET scans. From the results shown in \Cref{tab:ablation-n}, increasing the value of $N$ correlates with enhanced accuracy of the synthesized PET scans up to a certain threshold. However, beyond this threshold, further increments in $N$ lead to a decreased generation accuracy. This phenomenon may be attributed to the observation from \Cref{fig:ablation_N} that as $N$ increases, the applied strategy enhances the inter-slice consistency, albeit it induces a more pronounced smoothing effect on the generated scans. An optimal threshold for high-quality generation is identified at $N = 15$. For $N < 15$, the synthesized scans exhibit artifacts along the two additional axes (coronal and sagittal directions, given that we use axial slices as input), due to the slice inconsistency during generation. Conversely, for $N > 15$, the resulting scans are subject to excessive smoothing, leading to a loss of detailed information.

\begin{table}[t]
\small
 \centering
  {\setlength{\tabcolsep}{1.0em}
 \begin{tabular}{ccccc}\toprule
    $N$ & MAE \scriptsize{($\times 10^{-2}$)}$\downarrow$  & MSE \scriptsize{($\times 10^{-2}$)}$\downarrow$ & PSNR $\uparrow$ & SSIM(\%) $\uparrow$  \\\midrule
    1 & 4.05 & 0.60 & 23.10 & 83.09 \\
    5 & 3.64 & 0.50 & 24.07 & 85.47  \\
    11 & 3.55 & 0.46 & 24.33 & 85.79 \\
    15 & 3.45 & 0.43 & 24.59 & 86.29  \\
    19 & 3.59 & 0.45 & 24.26 & 84.79 
    \\\bottomrule
 \end{tabular}
 }
 \caption{Ablation on the number $N$ of input neighboring slices.}
 \label{tab:ablation-n}
\end{table}

\subsubsection{Directions of Input Slices \textcolor{modicolor}{and Cross-Axis Consistency}} 
\begin{figure}
    \centering
    \includegraphics[width=1\linewidth]{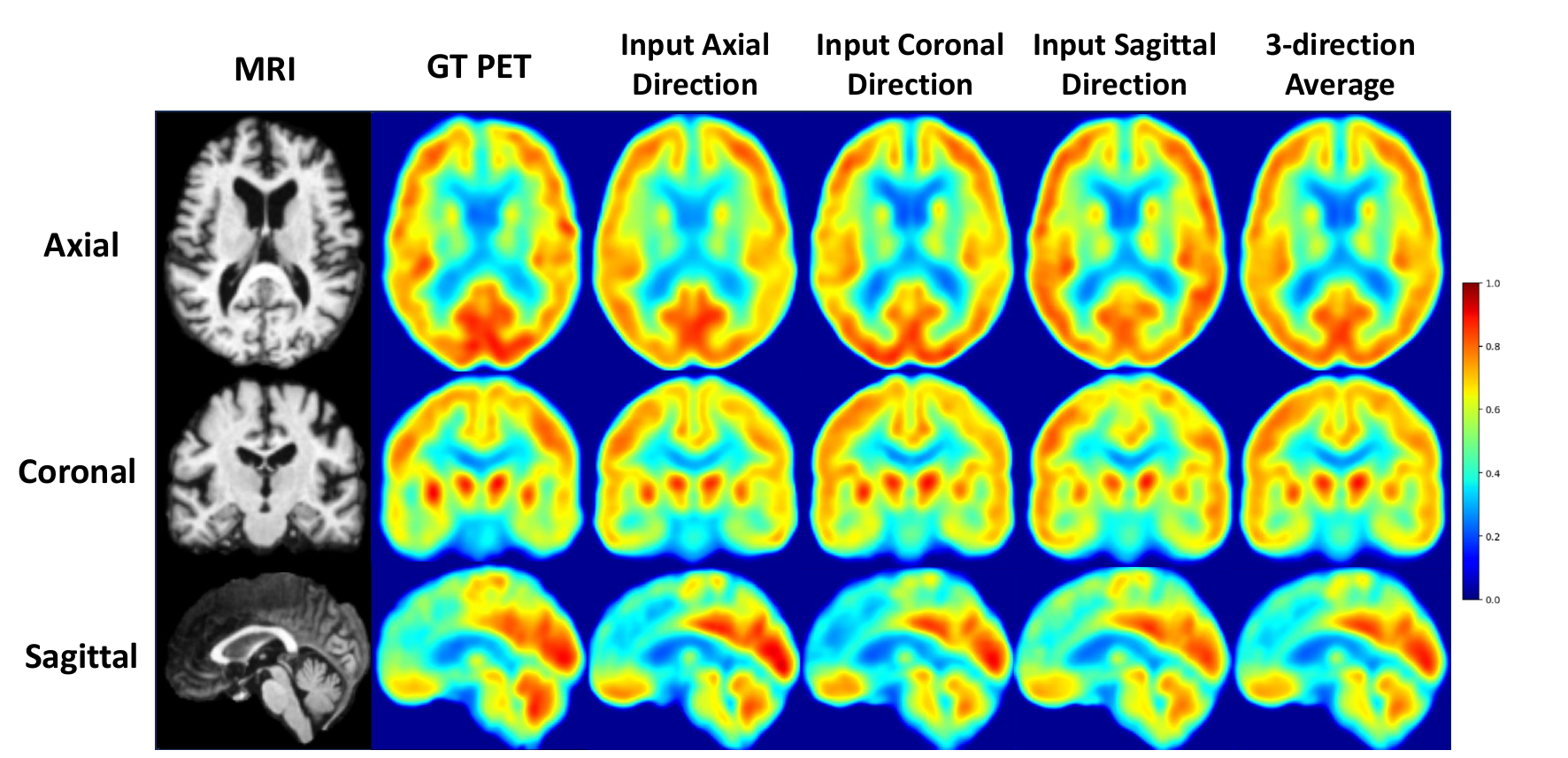}
    \caption{Ablation on the directions of input slices and the average of synthesized scans from the three directions. The influence of the input slice direction is minimal on the synthesized PET scans.}
    \label{fig:ablation_dir}
\end{figure}

Three-dimensional brain scans encompass three anatomical planes: axial, coronal, and sagittal planes. Our framework primarily takes slices from the axial plane as input. We make an assessment of the impact when receiving input slices from the coronal and sagittal planes instead. Furthermore, we average the synthesized scans from these three distinct networks, each trained on one of these planes, and compare the results qualitatively. \Cref{fig:ablation_dir} reveals that the anatomical direction of input slices exerts only slight effects on the quality of the synthesized PET. Furthermore, the 3-direction averaged scan also exhibits a minimal difference in quality. 
\textcolor{modicolor}{Quantitatively, reconstruction quality is consistent across all three orthogonal axes: taken the axial direction as input as an example, we obtain the image quality scores across three directions on the ADNI dataset as follows: SSIM (axial/coronal/sagittal) = 86.29\%/86.41\%/86.31\% and PSNR = 24.59/23.82/24.37 dB, respectively, indicating no axis-dependent degradation or cross-axis incoherence. 
\Cref{fig:ablation_dir} further shows minimal visual differences across planes and no artifacts or tearing; the 3-direction averaged scan is comparable to any single-axis model, reinforcing cross-axis spatial consistency.
Overall, the results demonstrate the robustness and spatial coherence of the proposed volumetric generation strategy in \modelName~across various slice directions. 
}

\subsection{Fairness Evaluation}

Fairness is crucial in medical imaging. It is essential that deep learning models employed in medical applications exhibit few biases towards specific demographic groups, including age, gender, or specific patient categories. 
To evaluate fairness, we assessed \modelName's performance across diverse patient cohorts with varying ages, genders, and diagnostic profiles on the test set.
The results presented in Table~\ref{tab:fairness} indicate that the errors of the synthesized PET across different groups exhibit minimal variance, with no statistically significant differences. Using the Wilcoxon rank-sum test and Bonferroni correction for multiple testing, all adjusted p-values exceed 0.05. 
These results indicate that \modelName~maintains consistent accuracy and performs equitably across all examined demographic categories.

\begin{table}[h]
\small
    \centering
    {\setlength{\tabcolsep}{1.5em}
    \begin{tabular}{ccc} \toprule
         Demographics & Groups & MAE ($\times 10^{-2}$) $ \downarrow$ \\\midrule
         \multirow{4}{*}{Age} & $<$ 60 & 3.55 \interval{0.55} \\
          & 60 - 70 & 3.30 \interval{0.36} \\
          & 70 - 80 & 3.44 \interval{0.50} \\
          & $>$ 80 & 3.66 \interval{0.67} \\\midrule
        \multirow{2}{*}{Gender} & Male & 3.52 \interval{0.57} \\
         & Female & 3.35 \interval{0.41} \\\midrule
        \multirow{3}{*}{Diagnosis} & CN & 3.26 \interval{0.44} \\
         & AD & 3.57 \interval{0.46} \\
         & MCI & 3.47 \interval{0.56} \\\midrule
        \multicolumn{2}{c}{Total} & 3.45 \interval{0.51} \\
       \bottomrule
        
    \end{tabular}}
    \caption{The mean absolute error (MAE) of synthesized PET on the test set among different demographics.}
    \label{tab:fairness}
\end{table}


\section{Conclusion}
\label{sec:conclusion}

We presented \modelName, a novel framework for brain MRI to PET translation leveraging conditional diffusion models.
\modelName~distinguishes itself from existing DM-based approaches by excelling in preserving both structural and pathological patterns in the target modality, which is enabled by its highly interactive dual-arm architecture with multi-modal conditioning.
The inclusion of CycleEx and a volumetric generation strategy further elevated its capability to produce high-quality 3D PET scans. 
Our comprehensive ablation study confirmed the effectiveness of these key designs.
In AD diagnosis, \modelName~achieved an improved AUC compared to real PET scans and surpasses MRI in diagnostic accuracy.  
Its distinct pathology awareness likely stems from the effective integration of multi-modal conditions and pathology priors, together with the synergy between its key components. 
\textcolor{modicolor}{Specifically, the incorporation of relevant clinical data alongside MRI enables the model to capture complementary cues that strengthen its sensitivity to disease-related changes. The use of MetaROIs as pathology priors further directs the model’s focus toward brain regions most affected by Alzheimer’s disease, such as key hypometabolic areas linked to abnormal metabolic activity. Finally, the CycleEx training strategy facilitates dual-arm information exchange, allowing the network to extract richer, task-specific features while preserving both structural integrity and pathological details during translation. These design choices collectively ensure that the generated PET images retain clinically meaningful pathological patterns, rather than merely reproducing structural information. 
A limitation of this study is the absence of a formal clinical reading study to quantitatively evaluate the diagnostic utility of the synthetic PET, which needs to be addressed in future work.}
Overall, \modelName~demonstrated high potential in bridging the gap between structural and functional brain degradation processes, offering promising advancements for clinical applications. \\


\textbf{Funding:} 
This work was supported by the German Research Foundation (DFG) and the Munich Center for Machine Learning (MCML). \\

\textbf{Acknowledgements:} 
We gratefully acknowledge the Leibniz Supercomputing Centre for providing the computing resources.



\clearpage

\appendix

\setcounter{table}{0}
\renewcommand{\thetable}{\Alph{section}.\arabic{table}}

\setcounter{figure}{0}  
\renewcommand{\thefigure}{\Alph{section}.\arabic{figure}}  

\color{modicolor}
\section{Computational Costs}
\label{sec:computational_costs}
We benchmarked the computational costs of different methods under identical hardware and training schedules (batch size $=6$). Results are summarized in Table~\ref{tab:comp_cost}. Among the baseline methods, BBDM requires 70~ms/step with 3.46~GB GPU memory, and BBDM-LDM requires 98~ms/step with 4.05~GB. ResViT incurs 308~ms/step with 6.94~GB. GAN-based models such as Pix2Pix and CycleGAN remain faster and lighter, whereas RegGAN requires higher time per step (214~ms) despite modest memory usage (3.78~GB). In this context, PASTA without CycleEx sits mid-pack in terms of speed and memory, being slower than BBDM and CycleGAN but faster than BBDM-LDM, with memory usage comparable to Pix2Pix. 

We then compared PASTA in the setup with and without CycleEx. Without CycleEx, PASTA requires 5.6~GB GPU memory and 82~ms/step. Incorporating CycleEx increases training step time by $\sim$3.68$\times$ (+268\%) and peak GPU memory by $\sim$3.29$\times$ (+229\%), yielding 302~ms/step and 18.40~GB memory usage. This overhead arises from the three additional forward/backward passes of the CycleEx pathways (without extra learnable parameters nor iterative sampling during training). Importantly, inference time is unaffected since CycleEx is only invoked during training as additional supervision. As shown in our ablation study (Table~5), CycleEx’s added computation translates into measurable accuracy improvements: MAE improves by more than 12\% (0.0345 with CycleEx vs.\ 0.0399 without CycleEx), albeit doubling the total training time (96~h vs.\ 48~h on the ADNI dataset), with zero inference penalty.

\begin{table}[ht]
\small
\centering
\color{modicolor}{
{\setlength{\tabcolsep}{0.8em}
\begin{tabular}{lcc}
\toprule
Method & GPU Memory (GB) & Time (ms/step) \\
\midrule
CycleGAN~\citep{CycleGAN}    & 4.70 & 67 \\
RegGAN~\citep{kong2021breaking}      & 3.78 & 214 \\
Pix2Pix~\citep{pix2pix}   & 5.02 & 25 \\
ResViT~\citep{resvit}     & 6.94 & 308 \\
BBDM-LDM~\citep{bbdm}    & 4.05 & 98 \\
BBDM~\citep{bbdm}       & 3.46 & 70 \\
\midrule
PASTA (Ours, w/o CycleEx) & 5.60 & 82 \\
PASTA (Ours, w/ CycleEx)  & 18.40 & 302 \\
\bottomrule
\end{tabular}}}
\caption{\textcolor{modicolor}{Computational cost of different methods under identical setup (batch size $=6$).}}
\label{tab:comp_cost}
\end{table}

\section{\textcolor{modicolor}{Slice-Aware Adaptive Group Normalization}}
\label{sec:sa-adagn}

We explored PASTA with Slice-Aware Adaptive Group Normalization (SA-AdaGN) to expose the diffusion model to through-plane position of the targeting generated slices. A normalized slice position $z\in[0,1]$ is first embedded using MLP and, analogously to the timestep, produces per-block scale and shift parameters that modulate the group normalization in every ResNet block across all UNet resolutions. The resulting conditional signal is combined with the existing conditioning (timestep $t$, task-specific representations $\boldsymbol{h_m}$, and clinical data $\boldsymbol{c}$) as input to the adaptive conditional module, resulting in an updated SA-AdaGN from Eq.~\ref{eq:adagn} as below:
\begin{align}
\label{eq:sa-adagn}
    \textrm {SA-AdaGN}(\boldsymbol{h}, t, \boldsymbol{c}, \boldsymbol{h_m}, z) =~& \boldsymbol{z_s}\boldsymbol{c_s}(\boldsymbol{h_m} (\boldsymbol{t_s}\textrm{GroupNorm}(\boldsymbol{h})+\boldsymbol{t_b}) + \boldsymbol{z_b}, \\ \nonumber
    (\boldsymbol{z_s}, \boldsymbol{z_b}) = ~&\text{MLP}(z).
\end{align}
Conceptually, SA-AdaGN aims to improve slice consistency in our proposed 2.5D volumetric generation strategy, by encoding the input slice position directly in the feature space. In practice, as shown in Table~\ref{tab:ablation-designs}, SA-AdaGN achieved performance broadly comparable to the original architecture, showing a slight increase in MAE but an improvement in SSIM, which points to a potential benefit in preserving structural information.
\textcolor{modicolor2}{
Furthermore, as shown in Table~\ref{tab:appendix_roi_metrics}, SA-AdaGN reached a slightly better performance with regards to the pathology-localized metrics, indicating enhanced fidelity within disease-relevant regions, even when global metrics show mixed behavior. 
Introducing the learnable weight factor $\lambda_R$ within the MetaROIs prior on top of SA-AdaGN provided only a marginal benefit, resulting in a slight improvement in the MAE in the ROI area ($8.76 \pm 3.14 \times 10^{-4}$ compared to $8.96 \pm 2.64 \times 10^{-4}$), with no improvement observed across the remaining quantitative metrics.
}
For simplicity, we therefore recommend the baseline PASTA configuration. Nevertheless, this presented architectural modification provides insights to inform future generative network design in medical imaging for incorporating additional through-plane positional priors.

\begin{table}[b]
\small
\centering
{\color{modicolor2}
{\setlength{\tabcolsep}{0.2em}
\begin{tabular}{lcccc}\toprule
Method 
& \scriptsize{MAE$_{ROI}$($\times 10^{-4}$)}$\downarrow$ 
& \scriptsize{MSE$_{ROI}$($\times 10^{-4}$)}$\downarrow$ 
& \scriptsize{PSNR$_{ROI}$$\uparrow$} 
& \scriptsize{SSIM$_{ROI}$(\%)}$\uparrow$ \\
\midrule

PASTA
    & 9.07 \interval{2.21}
    & 1.24 \interval{0.44}
    & 38.29 \interval{1.65}
    & 99.74 \interval{0.07} \\

PASTA \scriptsize{(CCL)}
    & 9.03 \interval{3.09}
    & 1.31 \interval{0.69}
    & 38.21 \interval{2.16}
    & 99.72 \interval{0.10} \\
PASTA \scriptsize{(CCL, $\lambda_{R}$ learn)}
    & 8.75 \interval{3.11}
    & 1.22 \interval{0.71}
    & 38.51 \interval{2.15}
    & 99.75 \interval{0.09} 
\\
PASTA \scriptsize{(SA-AdaGN)}
    & 8.96 \interval{2.64}
    & 1.17 \interval{0.50}
    & 38.64 \interval{1.75}
    & 99.76 \interval{0.06} \\

PASTA \scriptsize{(SA-AdaGN, $\lambda_R$ learn)}
    & 8.76 \interval{3.14}
    & 1.24 \interval{0.78}
    & 38.46 \interval{2.09}
    & 99.74 \interval{0.09} \\
\bottomrule
\end{tabular}}}
\caption{\textcolor{modicolor2}{Pathology-localized quantitative comparison on the ADNI dataset using ROI-based metrics computed within AD-related MetaROI regions. We report the mean $\pm$ standard deviation across the test set.}}
\label{tab:appendix_roi_metrics}
\end{table}

\section{Auxiliary Classifier Consistency Loss}
\label{sec:CCL}

We experimented augmenting the training process of PASTA with an in-loop auxiliary classifier that enforces disease label consistency on the translated PET images. Concretely, at each training step, the denoiser arm $\mathrm{x}_\theta$ predicts the clean PET scan $\hat{\mathcal{P}_0}$ from the noised input, which is fed to a ResNet-18 classifier \(f_{\phi}\) to obtain logits \(f_{\phi}(\hat{\mathcal{P}_0})\). We use a cross-entropy (CE) penalty with the ground-truth label \(y\) as an auxiliary classifier consistency loss as:
\begin{equation}
\mathcal{L}_{cls} \;=\; \mathrm{CE}~\!\big(f_{\phi}(\hat{\mathcal{P}_0}),\, y\big),
\end{equation}
resulting in the final overall training objective of PASTA as:
\begin{equation}
    \mathcal{L} = \lambda_{task} * \mathcal{L}_{task} + \lambda_{diff} * \mathcal{L}_{diff} + \lambda_{cycle} * \mathcal{L}_{cycle} + \lambda_{cls}\ * \mathcal{L}_{cls}, 
\end{equation}
where \(\lambda_{cls}\) is a constant variable controlling the contribution of this auxiliary term, where we set $\lambda_{cls}=0.001$ after the hyperparameter search. The classifier \(f_{\phi}\) can be initialized either from pretrained weights (optionally frozen during training) or from random initialization and trained jointly with the diffusion model. We evaluated these alternatives and observed that training the classifier from scratch achieved performance comparable to the pretrained variants. For simplicity and reproducibility, we therefore adopted the joint training from scratch as our default configuration. To provide a stable supervision signal, we activate the auxiliary classifier consistency loss after a brief warm-up period, once the generated PET images are sufficiently realistic for reliable classification. 
This loss term addition aims to couple a discriminative signal in the generative process, which can guide the synthesis toward pathology-aware, semantically faithful outputs and may aid interpretability. Empirically, as shown in the ablation results Table~\ref{tab:ablation-designs}, the added classifier consistency loss (CCL) yielded comparable results to our baseline PASTA setup, with a slight increase in pixel-wise error but an improvement in SSIM, suggesting a potential advantage in preserving structural fidelity. 
\textcolor{modicolor2}{In the pathology-localized metrics, as shown in Table~\ref{tab:appendix_roi_metrics}, CCL also reached comparable results to the baseline PASTA, with a slightly better performance in MAE$_{ROI}$.
Incorporating the learnable weight factor $\lambda_R$ within the MetaROIs prior on top of the CCL module offered a marginal improvement, leading to a slight improvement in the MAE in the ROI area ($8.75 \pm 3.11 \times 10^{-4}$ compared to $8.96 \pm 2.64 \times 10^{-4}$), with no improvement observed across the remaining quantitative metrics.
}
We therefore recommend the original setup for simplicity. However, this modification is promising for future studies in further fostering pathology awareness and interpretability with discriminative learning during medical image translation.

\color{black}









\clearpage

\bibliographystyle{elsarticle-num}

\begin{thebibliography}{46}
\expandafter\ifx\csname natexlab\endcsname\relax\def\natexlab#1{#1}\fi
\providecommand{\url}[1]{\texttt{#1}}
\providecommand{\href}[2]{#2}
\providecommand{\path}[1]{#1}
\providecommand{\DOIprefix}{doi:}
\providecommand{\ArXivprefix}{arXiv:}
\providecommand{\URLprefix}{URL: }
\providecommand{\Pubmedprefix}{pmid:}
\providecommand{\doi}[1]{\href{http://dx.doi.org/#1}{\path{#1}}}
\providecommand{\Pubmed}[1]{\href{pmid:#1}{\path{#1}}}
\providecommand{\bibinfo}[2]{#2}
\ifx\xfnm\relax \def\xfnm[#1]{\unskip,\space#1}\fi
\bibitem[{Aisen et~al.(2017)Aisen, Cummings, Jack, Morris, Sperling, Fr{\"o}lich, Jones, Dowsett, Matthews, Raskin, Scheltens, and Dubois}]{addiagnosis}
\bibinfo{author}{P.~S. Aisen}, \bibinfo{author}{J.~Cummings}, \bibinfo{author}{C.~R. Jack}, \bibinfo{author}{J.~C. Morris}, \bibinfo{author}{R.~Sperling}, \bibinfo{author}{L.~Fr{\"o}lich}, \bibinfo{author}{R.~W. Jones}, \bibinfo{author}{S.~A. Dowsett}, \bibinfo{author}{B.~R. Matthews}, \bibinfo{author}{J.~Raskin}, \bibinfo{author}{P.~Scheltens}, \bibinfo{author}{B.~Dubois},
\newblock \bibinfo{title}{On the path to 2025: understanding the alzheimer’s disease continuum},
\newblock \bibinfo{journal}{Alzheimer's research \& therapy} \bibinfo{volume}{9} (\bibinfo{year}{2017}) \bibinfo{pages}{1--10}.
\bibitem[{Chouliaras and O'Brien(2023)}]{mri_pet}
\bibinfo{author}{L.~Chouliaras}, \bibinfo{author}{J.~O'Brien},
\newblock \bibinfo{title}{The use of neuroimaging techniques in the early and differential diagnosis of dementia},
\newblock \bibinfo{journal}{Molecular psychiatry}  (\bibinfo{year}{2023}).
\bibitem[{Kyrtata et~al.(2021)Kyrtata, Emsley, Sparasci, Parkes, and Dickie}]{ad_glucose}
\bibinfo{author}{N.~Kyrtata}, \bibinfo{author}{H.~C. Emsley}, \bibinfo{author}{O.~Sparasci}, \bibinfo{author}{L.~M. Parkes}, \bibinfo{author}{B.~R. Dickie},
\newblock \bibinfo{title}{A systematic review of glucose transport alterations in alzheimer's disease},
\newblock \bibinfo{journal}{Frontiers in Neuroscience} \bibinfo{volume}{15} (\bibinfo{year}{2021}) \bibinfo{pages}{626636}.
\bibitem[{Marcus et~al.(2014)Marcus, Mena, and Subramaniam}]{pet_differential}
\bibinfo{author}{C.~Marcus}, \bibinfo{author}{E.~Mena}, \bibinfo{author}{R.~M. Subramaniam},
\newblock \bibinfo{title}{Brain pet in the diagnosis of alzheimer’s disease},
\newblock \bibinfo{journal}{Clinical nuclear medicine} \bibinfo{volume}{39} (\bibinfo{year}{2014}) \bibinfo{pages}{e413}.
\bibitem[{Bloudek et~al.(2011)Bloudek, Spackman, Blankenburg, and Sullivan}]{pet_acc}
\bibinfo{author}{L.~M. Bloudek}, \bibinfo{author}{D.~E. Spackman}, \bibinfo{author}{M.~Blankenburg}, \bibinfo{author}{S.~D. Sullivan},
\newblock \bibinfo{title}{Review and meta-analysis of biomarkers and diagnostic imaging in alzheimer's disease},
\newblock \bibinfo{journal}{Journal of Alzheimer's Disease} \bibinfo{volume}{26} (\bibinfo{year}{2011}) \bibinfo{pages}{627--645}.
\bibitem[{Keppler and Conti(2001)}]{pet_cost}
\bibinfo{author}{J.~S. Keppler}, \bibinfo{author}{P.~S. Conti},
\newblock \bibinfo{title}{A cost analysis of positron emission tomography},
\newblock \bibinfo{journal}{American Journal of Roentgenology} \bibinfo{volume}{177} (\bibinfo{year}{2001}) \bibinfo{pages}{31--40}.
\bibitem[{Frisoni et~al.(2013)Frisoni, Bocchetta, Chételat, Rabinovici, de~Leon, Kaye, Reiman, Scheltens, Barkhof, Black, Brooks, Carrillo, Fox, Herholz, Nordberg, Jack, Jagust, Johnson, Rowe, Sperling, Thies, Wahlund, Weiner, Pasqualetti, DeCarli, and Area}]{pet_ov_mri}
\bibinfo{author}{G.~B. Frisoni}, \bibinfo{author}{M.~Bocchetta}, \bibinfo{author}{G.~Chételat}, \bibinfo{author}{G.~D. Rabinovici}, \bibinfo{author}{M.~J. de~Leon}, \bibinfo{author}{J.~Kaye}, \bibinfo{author}{E.~M. Reiman}, \bibinfo{author}{P.~Scheltens}, \bibinfo{author}{F.~Barkhof}, \bibinfo{author}{S.~E. Black}, \bibinfo{author}{D.~J. Brooks}, \bibinfo{author}{M.~C. Carrillo}, \bibinfo{author}{N.~C. Fox}, \bibinfo{author}{K.~Herholz}, \bibinfo{author}{A.~Nordberg}, \bibinfo{author}{C.~R. Jack}, \bibinfo{author}{W.~J. Jagust}, \bibinfo{author}{K.~A. Johnson}, \bibinfo{author}{C.~C. Rowe}, \bibinfo{author}{R.~A. Sperling}, \bibinfo{author}{W.~Thies}, \bibinfo{author}{L.-O. Wahlund}, \bibinfo{author}{M.~W. Weiner}, \bibinfo{author}{P.~Pasqualetti}, \bibinfo{author}{C.~DeCarli}, \bibinfo{author}{F.~I. N. P.~I. Area},
\newblock \bibinfo{title}{Imaging markers for alzheimer disease: which vs how},
\newblock \bibinfo{journal}{Neurology} \bibinfo{volume}{81} (\bibinfo{year}{2013}) \bibinfo{pages}{487--500}.
\bibitem[{Ho et~al.(2020)Ho, Jain, and Abbeel}]{ddpm}
\bibinfo{author}{J.~Ho}, \bibinfo{author}{A.~Jain}, \bibinfo{author}{P.~Abbeel},
\newblock \bibinfo{title}{Denoising diffusion probabilistic models},
\newblock in: \bibinfo{booktitle}{NeurIPS}, \bibinfo{year}{2020}.
\bibitem[{Dhariwal and Nichol(2021)}]{DMbeatsGAN}
\bibinfo{author}{P.~Dhariwal}, \bibinfo{author}{A.~Nichol},
\newblock \bibinfo{title}{Diffusion models beat gans on image synthesis},
\newblock in: \bibinfo{booktitle}{NeurIPS}, \bibinfo{year}{2021}.
\bibitem[{Li et~al.(2023)Li, Xue, Liu, and Lai}]{bbdm}
\bibinfo{author}{B.~Li}, \bibinfo{author}{K.~Xue}, \bibinfo{author}{B.~Liu}, \bibinfo{author}{Y.~Lai},
\newblock \bibinfo{title}{Bbdm: Image-to-image translation with brownian bridge diffusion models},
\newblock in: \bibinfo{booktitle}{CVPR}, \bibinfo{year}{2023}.
\bibitem[{Saharia et~al.(2022)Saharia, Chan, Chang, Lee, Ho, Salimans, Fleet, and Norouzi}]{Palette}
\bibinfo{author}{C.~Saharia}, \bibinfo{author}{W.~Chan}, \bibinfo{author}{H.~Chang}, \bibinfo{author}{C.~A. Lee}, \bibinfo{author}{J.~Ho}, \bibinfo{author}{T.~Salimans}, \bibinfo{author}{D.~J. Fleet}, \bibinfo{author}{M.~Norouzi},
\newblock \bibinfo{title}{Palette: Image-to-image diffusion models},
\newblock in: \bibinfo{booktitle}{ACM SIGGRAPH}, \bibinfo{year}{2022}.
\bibitem[{Wei et~al.(2018)Wei, Poirion, Bodini, Durrleman, Ayache, Stankoff, and Colliot}]{sketcher-refine}
\bibinfo{author}{W.~Wei}, \bibinfo{author}{E.~Poirion}, \bibinfo{author}{B.~Bodini}, \bibinfo{author}{S.~Durrleman}, \bibinfo{author}{N.~Ayache}, \bibinfo{author}{B.~Stankoff}, \bibinfo{author}{O.~Colliot},
\newblock \bibinfo{title}{Learning myelin content in multiple sclerosis from multimodal mri through adversarial training},
\newblock in: \bibinfo{booktitle}{MICCAI}, \bibinfo{year}{2018}.
\bibitem[{Lan et~al.(2021)Lan, Initiative, Toga, and Sepehrband}]{SCGAN}
\bibinfo{author}{H.~Lan}, \bibinfo{author}{A.~D.~N. Initiative}, \bibinfo{author}{A.~W. Toga}, \bibinfo{author}{F.~Sepehrband},
\newblock \bibinfo{title}{Three-dimensional self-attention conditional gan with spectral normalization for multimodal neuroimaging synthesis},
\newblock \bibinfo{journal}{Magnetic resonance in medicine} \bibinfo{volume}{86} (\bibinfo{year}{2021}) \bibinfo{pages}{1718--1733}.
\bibitem[{Zhang et~al.(2022)Zhang, He, Qing, Gao, and Wang}]{bpgan}
\bibinfo{author}{J.~Zhang}, \bibinfo{author}{X.~He}, \bibinfo{author}{L.~Qing}, \bibinfo{author}{F.~Gao}, \bibinfo{author}{B.~Wang},
\newblock \bibinfo{title}{Bpgan: Brain pet synthesis from mri using generative adversarial network for multi-modal alzheimer’s disease diagnosis},
\newblock \bibinfo{journal}{Computer Methods and Programs in Biomedicine} \bibinfo{volume}{217} (\bibinfo{year}{2022}) \bibinfo{pages}{106676}.
\bibitem[{Hu et~al.(2021)Hu, Lei, Wang, Wang, Feng, and Shen}]{bidirectionalGAN}
\bibinfo{author}{S.~Hu}, \bibinfo{author}{B.~Lei}, \bibinfo{author}{S.~Wang}, \bibinfo{author}{Y.~Wang}, \bibinfo{author}{Z.~Feng}, \bibinfo{author}{Y.~Shen},
\newblock \bibinfo{title}{Bidirectional mapping generative adversarial networks for brain mr to pet synthesis},
\newblock \bibinfo{journal}{IEEE Transactions on Medical Imaging} \bibinfo{volume}{41} (\bibinfo{year}{2021}) \bibinfo{pages}{145--157}.
\bibitem[{Shin et~al.(2020)Shin, Ihsani, Mandava, Sreenivas, Forster, Cha, and Initiative}]{ganbert}
\bibinfo{author}{H.-C. Shin}, \bibinfo{author}{A.~Ihsani}, \bibinfo{author}{S.~Mandava}, \bibinfo{author}{S.~T. Sreenivas}, \bibinfo{author}{C.~Forster}, \bibinfo{author}{J.~Cha}, \bibinfo{author}{A.~D.~N. Initiative},
\newblock \bibinfo{title}{Ganbert: Generative adversarial networks with bidirectional encoder representations from transformers for mri to pet synthesis},
\newblock \bibinfo{journal}{arXiv:2008.04393}  (\bibinfo{year}{2020}).
\bibitem[{Dalmaz et~al.(2022)Dalmaz, Yurt, and Çukur}]{resvit}
\bibinfo{author}{O.~Dalmaz}, \bibinfo{author}{M.~Yurt}, \bibinfo{author}{T.~Çukur},
\newblock \bibinfo{title}{Resvit: Residual vision transformers for multimodal medical image synthesis},
\newblock \bibinfo{journal}{IEEE Transactions on Medical Imaging} \bibinfo{volume}{41} (\bibinfo{year}{2022}) \bibinfo{pages}{2598--2614}.
\bibitem[{Shin et~al.(2020)Shin, Ihsani, Xu, Mandava, Sreenivas, Forster, and Cha}]{gandalf}
\bibinfo{author}{H.-C. Shin}, \bibinfo{author}{A.~Ihsani}, \bibinfo{author}{Z.~Xu}, \bibinfo{author}{S.~Mandava}, \bibinfo{author}{S.~T. Sreenivas}, \bibinfo{author}{C.~Forster}, \bibinfo{author}{J.~Cha},
\newblock \bibinfo{title}{Gandalf: Generative adversarial networks with discriminator-adaptive loss fine-tuning for alzheimer’s disease diagnosis from mri},
\newblock in: \bibinfo{booktitle}{MICCAI}, \bibinfo{year}{2020}.
\bibitem[{Sohl-Dickstein et~al.(2015)Sohl-Dickstein, Weiss, Maheswaranathan, and Ganguli}]{diffusion}
\bibinfo{author}{J.~Sohl-Dickstein}, \bibinfo{author}{E.~Weiss}, \bibinfo{author}{N.~Maheswaranathan}, \bibinfo{author}{S.~Ganguli},
\newblock \bibinfo{title}{Deep unsupervised learning using nonequilibrium thermodynamics},
\newblock in: \bibinfo{booktitle}{ICML}, \bibinfo{year}{2015}.
\bibitem[{Zhu et~al.(2023)Zhu, Xue, Jin, Liu, He, Liu, and Yu}]{make-a-volume}
\bibinfo{author}{L.~Zhu}, \bibinfo{author}{Z.~Xue}, \bibinfo{author}{Z.~Jin}, \bibinfo{author}{X.~Liu}, \bibinfo{author}{J.~He}, \bibinfo{author}{Z.~Liu}, \bibinfo{author}{L.~Yu},
\newblock \bibinfo{title}{Make-a-volume: Leveraging latent diffusion models for cross-modality 3d brain mri synthesis},
\newblock in: \bibinfo{booktitle}{MICCAI}, \bibinfo{year}{2023}.
\bibitem[{Peng et~al.(2023)Peng, Adeli, Bosschieter, Park, Zhao, and Pohl}]{generate_real_MRI_cDPM}
\bibinfo{author}{W.~Peng}, \bibinfo{author}{E.~Adeli}, \bibinfo{author}{T.~Bosschieter}, \bibinfo{author}{S.~H. Park}, \bibinfo{author}{Q.~Zhao}, \bibinfo{author}{K.~M. Pohl},
\newblock \bibinfo{title}{Generating realistic brain mris via a conditional diffusion probabilistic model},
\newblock in: \bibinfo{booktitle}{MICCAI}, \bibinfo{year}{2023}.
\bibitem[{{\"O}zbey et~al.(2023){\"O}zbey, Dalmaz, Dar, Bedel, {\"O}zturk, G{\"u}ng{\"o}r, and {\c{C}}ukur}]{syndiff}
\bibinfo{author}{M.~{\"O}zbey}, \bibinfo{author}{O.~Dalmaz}, \bibinfo{author}{S.~U. Dar}, \bibinfo{author}{H.~A. Bedel}, \bibinfo{author}{{\c{S}}.~{\"O}zturk}, \bibinfo{author}{A.~G{\"u}ng{\"o}r}, \bibinfo{author}{T.~{\c{C}}ukur},
\newblock \bibinfo{title}{Unsupervised medical image translation with adversarial diffusion models},
\newblock \bibinfo{journal}{IEEE Transactions on Medical Imaging}  (\bibinfo{year}{2023}).
\bibitem[{Xie et~al.(2024)Xie, Cao, Cui, Guo, Wu, Wang, Li, Hu, Sun, Sang, Zhou, Zhu, Liang, Jin, Zeng, Chen, and Wang}]{xie2024synpet}
\bibinfo{author}{T.~Xie}, \bibinfo{author}{C.~Cao}, \bibinfo{author}{Z.-x. Cui}, \bibinfo{author}{Y.~Guo}, \bibinfo{author}{C.~Wu}, \bibinfo{author}{X.~Wang}, \bibinfo{author}{Q.~Li}, \bibinfo{author}{Z.~Hu}, \bibinfo{author}{T.~Sun}, \bibinfo{author}{Z.~Sang}, \bibinfo{author}{Y.~Zhou}, \bibinfo{author}{Y.~Zhu}, \bibinfo{author}{D.~Liang}, \bibinfo{author}{Q.~Jin}, \bibinfo{author}{H.~Zeng}, \bibinfo{author}{G.~Chen}, \bibinfo{author}{H.~Wang},
\newblock \bibinfo{title}{Synthesizing pet images from high-field and ultra-high-field mr images using joint diffusion attention model},
\newblock \bibinfo{journal}{Medical Physics} \bibinfo{volume}{51} (\bibinfo{year}{2024}) \bibinfo{pages}{5250--5269}. \DOIprefix\doi{https://doi.org/10.1002/mp.17254}.
\bibitem[{Li et~al.(2024)Li, Yakushev, Hedderich, and Wachinger}]{Li2024pasta}
\bibinfo{author}{Y.~Li}, \bibinfo{author}{I.~Yakushev}, \bibinfo{author}{D.~M. Hedderich}, \bibinfo{author}{C.~Wachinger},
\newblock \bibinfo{title}{Pasta: Pathology-aware mri to pet cross-modal translation with diffusion models},
\newblock in: \bibinfo{booktitle}{Medical Image Computing and Computer Assisted Intervention -- MICCAI 2024}, \bibinfo{publisher}{Springer Nature Switzerland}, \bibinfo{address}{Cham}, \bibinfo{year}{2024}, pp. \bibinfo{pages}{529--540}.
\bibitem[{Su et~al.(2023)Su, Song, Meng, and Ermon}]{ddib}
\bibinfo{author}{X.~Su}, \bibinfo{author}{J.~Song}, \bibinfo{author}{C.~Meng}, \bibinfo{author}{S.~Ermon},
\newblock \bibinfo{title}{Dual diffusion implicit bridges for image-to-image translation},
\newblock in: \bibinfo{booktitle}{ICLR}, \bibinfo{year}{2023}.
\bibitem[{Kwon and Ye(2023)}]{diffuseIT}
\bibinfo{author}{G.~Kwon}, \bibinfo{author}{J.~C. Ye},
\newblock \bibinfo{title}{Diffusion-based image translation using disentangled style and content representation},
\newblock in: \bibinfo{booktitle}{ICLR}, \bibinfo{year}{2023}.
\bibitem[{Nichol and Dhariwal(2021)}]{improved_ddpm}
\bibinfo{author}{A.~Q. Nichol}, \bibinfo{author}{P.~Dhariwal},
\newblock \bibinfo{title}{Improved denoising diffusion probabilistic models},
\newblock in: \bibinfo{booktitle}{ICML}, \bibinfo{year}{2021}.
\bibitem[{Rombach et~al.(2022)Rombach, Blattmann, Lorenz, Esser, and Ommer}]{stable_diffusion}
\bibinfo{author}{R.~Rombach}, \bibinfo{author}{A.~Blattmann}, \bibinfo{author}{D.~Lorenz}, \bibinfo{author}{P.~Esser}, \bibinfo{author}{B.~Ommer},
\newblock \bibinfo{title}{High-resolution image synthesis with latent diffusion models},
\newblock in: \bibinfo{booktitle}{CVPR}, \bibinfo{year}{2022}.
\bibitem[{Salimans and Ho(2022)}]{progressive}
\bibinfo{author}{T.~Salimans}, \bibinfo{author}{J.~Ho},
\newblock \bibinfo{title}{Progressive distillation for fast sampling of diffusion models},
\newblock in: \bibinfo{booktitle}{ICLR}, \bibinfo{year}{2022}.
\bibitem[{Ronneberger et~al.(2015)Ronneberger, Fischer, and Brox}]{unet}
\bibinfo{author}{O.~Ronneberger}, \bibinfo{author}{P.~Fischer}, \bibinfo{author}{T.~Brox},
\newblock \bibinfo{title}{U-net: Convolutional networks for biomedical image segmentation},
\newblock in: \bibinfo{booktitle}{MICCAI}, \bibinfo{year}{2015}.
\bibitem[{Folstein et~al.(1975)Folstein, Folstein, and McHugh}]{mmse}
\bibinfo{author}{M.~F. Folstein}, \bibinfo{author}{S.~E. Folstein}, \bibinfo{author}{P.~R. McHugh},
\newblock \bibinfo{title}{“mini-mental state”: a practical method for grading the cognitive state of patients for the clinician},
\newblock \bibinfo{journal}{Journal of psychiatric research} \bibinfo{volume}{12} (\bibinfo{year}{1975}) \bibinfo{pages}{189--198}.
\bibitem[{Mohs et~al.(1997)Mohs, Knopman, Petersen, Ferris, Ernesto, Grundman, Sano, Bieliauskas, Geldmacher, Clark, and Thai}]{adas-cog-13}
\bibinfo{author}{R.~C. Mohs}, \bibinfo{author}{D.~Knopman}, \bibinfo{author}{R.~C. Petersen}, \bibinfo{author}{S.~H. Ferris}, \bibinfo{author}{C.~Ernesto}, \bibinfo{author}{M.~Grundman}, \bibinfo{author}{M.~Sano}, \bibinfo{author}{L.~Bieliauskas}, \bibinfo{author}{D.~Geldmacher}, \bibinfo{author}{C.~Clark}, \bibinfo{author}{L.~J. Thai},
\newblock \bibinfo{title}{Development of cognitive instruments for use in clinical trials of antidementia drugs: additions to the alzheimer's disease assessment scale that broaden its scope},
\newblock \bibinfo{journal}{Alzheimer Disease \& Associated Disorders} \bibinfo{volume}{11} (\bibinfo{year}{1997}) \bibinfo{pages}{13--21}.
\bibitem[{Strittmatter and Roses(1996)}]{apoe4}
\bibinfo{author}{W.~J. Strittmatter}, \bibinfo{author}{A.~D. Roses},
\newblock \bibinfo{title}{Apolipoprotein e and alzheimer's disease},
\newblock \bibinfo{journal}{Annual review of neuroscience} \bibinfo{volume}{19} (\bibinfo{year}{1996}) \bibinfo{pages}{53--77}.
\bibitem[{Jarrett et~al.(2019)Jarrett, Yoon, and van~der Schaar}]{tabularselect}
\bibinfo{author}{D.~Jarrett}, \bibinfo{author}{J.~Yoon}, \bibinfo{author}{M.~van~der Schaar},
\newblock \bibinfo{title}{Dynamic prediction in clinical survival analysis using temporal convolutional networks},
\newblock \bibinfo{journal}{IEEE journal of biomedical and health informatics} \bibinfo{volume}{24} (\bibinfo{year}{2019}) \bibinfo{pages}{424--436}.
\bibitem[{Landau et~al.(2011)Landau, Harvey, Madison, Koeppe, Reiman, Foster, Weiner, and Jagust}]{metaROIs}
\bibinfo{author}{S.~M. Landau}, \bibinfo{author}{D.~Harvey}, \bibinfo{author}{C.~M. Madison}, \bibinfo{author}{R.~A. Koeppe}, \bibinfo{author}{E.~M. Reiman}, \bibinfo{author}{N.~L. Foster}, \bibinfo{author}{M.~W. Weiner}, \bibinfo{author}{W.~J. Jagust},
\newblock \bibinfo{title}{Associations between cognitive, functional, and fdg-pet measures of decline in ad and mci},
\newblock \bibinfo{journal}{Neurobiology of Aging} \bibinfo{volume}{32} (\bibinfo{year}{2011}) \bibinfo{pages}{1207--1218}.
\bibitem[{Zhu et~al.(2017)Zhu, Park, Isola, and Efros}]{CycleGAN}
\bibinfo{author}{J.-Y. Zhu}, \bibinfo{author}{T.~Park}, \bibinfo{author}{P.~Isola}, \bibinfo{author}{A.~A. Efros},
\newblock \bibinfo{title}{Unpaired image-to-image translation using cycle-consistent adversarial networkss},
\newblock in: \bibinfo{booktitle}{ICCV}, \bibinfo{year}{2017}.
\bibitem[{Jack~Jr. et~al.(2008)Jack~Jr., Bernstein, Fox, Thompson, Alexander, Harvey, Borowski, Britson, L.~Whitwell, Ward, Dale, Felmlee, Gunter, Hill, Killiany, Schuff, Fox-Bosetti, Lin, Studholme, DeCarli, Krueger, Ward, Metzger, Scott, Mallozzi, Blezek, Levy, Debbins, Fleisher, Albert, Green, Bartzokis, Glover, Mugler, and Weiner}]{adni}
\bibinfo{author}{C.~R. Jack~Jr.}, \bibinfo{author}{M.~A. Bernstein}, \bibinfo{author}{N.~C. Fox}, \bibinfo{author}{P.~Thompson}, \bibinfo{author}{G.~Alexander}, \bibinfo{author}{D.~Harvey}, \bibinfo{author}{B.~Borowski}, \bibinfo{author}{P.~J. Britson}, \bibinfo{author}{J.~L.~Whitwell}, \bibinfo{author}{C.~Ward}, \bibinfo{author}{A.~M. Dale}, \bibinfo{author}{J.~P. Felmlee}, \bibinfo{author}{J.~L. Gunter}, \bibinfo{author}{D.~L. Hill}, \bibinfo{author}{R.~Killiany}, \bibinfo{author}{N.~Schuff}, \bibinfo{author}{S.~Fox-Bosetti}, \bibinfo{author}{C.~Lin}, \bibinfo{author}{C.~Studholme}, \bibinfo{author}{C.~S. DeCarli}, \bibinfo{author}{G.~Krueger}, \bibinfo{author}{H.~A. Ward}, \bibinfo{author}{G.~J. Metzger}, \bibinfo{author}{K.~T. Scott}, \bibinfo{author}{R.~Mallozzi}, \bibinfo{author}{D.~Blezek}, \bibinfo{author}{J.~Levy}, \bibinfo{author}{J.~P. Debbins}, \bibinfo{author}{A.~S. Fleisher}, \bibinfo{author}{M.~Albert}, \bibinfo{author}{R.~Green}, \bibinfo{author}{G.~Bartzokis}, \bibinfo{author}{G.~Glover},
  \bibinfo{author}{J.~Mugler}, \bibinfo{author}{M.~W. Weiner},
\newblock \bibinfo{title}{The alzheimer's disease neuroimaging initiative (adni): Mri methods},
\newblock \bibinfo{journal}{Journal of Magnetic Resonance Imaging} \bibinfo{volume}{27} (\bibinfo{year}{2008}) \bibinfo{pages}{685--691}.
\bibitem[{Hoopes et~al.(2022)Hoopes, Mora, Dalca, Fischl, and Hoffmann}]{synthstrip}
\bibinfo{author}{A.~Hoopes}, \bibinfo{author}{J.~S. Mora}, \bibinfo{author}{A.~V. Dalca}, \bibinfo{author}{B.~Fischl}, \bibinfo{author}{M.~Hoffmann},
\newblock \bibinfo{title}{Synthstrip: skull-stripping for any brain image},
\newblock \bibinfo{journal}{NeuroImage} \bibinfo{volume}{260} (\bibinfo{year}{2022}) \bibinfo{pages}{119474}.
\bibitem[{Fischl(2012)}]{fischlFreeSurfer2012}
\bibinfo{author}{B.~Fischl},
\newblock \bibinfo{title}{{{FreeSurfer}}},
\newblock \bibinfo{journal}{NeuroImage} \bibinfo{volume}{62} (\bibinfo{year}{2012}) \bibinfo{pages}{774--781}.
\bibitem[{Thibeau-Sutre et~al.(2022)Thibeau-Sutre, Diaz, Hassanaly, Routier, Dormont, Colliot, and Burgos}]{clinicadl}
\bibinfo{author}{E.~Thibeau-Sutre}, \bibinfo{author}{M.~Diaz}, \bibinfo{author}{R.~Hassanaly}, \bibinfo{author}{A.~Routier}, \bibinfo{author}{D.~Dormont}, \bibinfo{author}{O.~Colliot}, \bibinfo{author}{N.~Burgos},
\newblock \bibinfo{title}{{ClinicaDL: an open-source deep learning software for reproducible neuroimaging processing}},
\newblock \bibinfo{journal}{{Computer Methods and Programs in Biomedicine}} \bibinfo{volume}{220} (\bibinfo{year}{2022}) \bibinfo{pages}{106818}. \DOIprefix\doi{10.1016/j.cmpb.2022.106818}.
\bibitem[{Ho et~al.(2007)Ho, Imai, King, and Stuart}]{ho2007matching}
\bibinfo{author}{D.~E. Ho}, \bibinfo{author}{K.~Imai}, \bibinfo{author}{G.~King}, \bibinfo{author}{E.~A. Stuart},
\newblock \bibinfo{title}{Matching as nonparametric preprocessing for reducing model dependence in parametric causal inference},
\newblock \bibinfo{journal}{Political analysis} \bibinfo{volume}{15} (\bibinfo{year}{2007}) \bibinfo{pages}{199--236}.
\bibitem[{Song et~al.(2021)Song, Meng, and Ermon}]{ddim}
\bibinfo{author}{J.~Song}, \bibinfo{author}{C.~Meng}, \bibinfo{author}{S.~Ermon},
\newblock \bibinfo{title}{Denoising diffusion implicit models},
\newblock in: \bibinfo{booktitle}{ICLR}, \bibinfo{year}{2021}.
\bibitem[{Isola et~al.(2017)Isola, Zhu, Zhou, and Efros}]{pix2pix}
\bibinfo{author}{P.~Isola}, \bibinfo{author}{J.-Y. Zhu}, \bibinfo{author}{T.~Zhou}, \bibinfo{author}{A.~A. Efros},
\newblock \bibinfo{title}{Image-to-image translation with conditional adversarial networks},
\newblock in: \bibinfo{booktitle}{CVPR}, \bibinfo{year}{2017}.
\bibitem[{Hoffman et~al.(2000)Hoffman, Welsh-Bohmer, Hanson, Crain, Hulette, Earl, and Coleman}]{pet-hypometa}
\bibinfo{author}{J.~M. Hoffman}, \bibinfo{author}{K.~A. Welsh-Bohmer}, \bibinfo{author}{M.~Hanson}, \bibinfo{author}{B.~Crain}, \bibinfo{author}{C.~Hulette}, \bibinfo{author}{N.~Earl}, \bibinfo{author}{R.~E. Coleman},
\newblock \bibinfo{title}{Fdg pet imaging in patients with pathologically verified dementia},
\newblock \bibinfo{journal}{Journal of Nuclear Medicine} \bibinfo{volume}{41} (\bibinfo{year}{2000}) \bibinfo{pages}{1920--1928}.
\bibitem[{Minoshima et~al.(1995)Minoshima, Frey, Koeppe, Foster, and Kuhl}]{3dssp}
\bibinfo{author}{S.~Minoshima}, \bibinfo{author}{K.~A. Frey}, \bibinfo{author}{R.~A. Koeppe}, \bibinfo{author}{N.~L. Foster}, \bibinfo{author}{D.~E. Kuhl},
\newblock \bibinfo{title}{A diagnostic approach in alzheimer's disease using three-dimensional stereotactic surface projections of fluorine-18-fdg pet},
\newblock \bibinfo{journal}{Journal of Nuclear Medicine} \bibinfo{volume}{36} (\bibinfo{year}{1995}) \bibinfo{pages}{1238--1248}.
\bibitem[{Herscovitch(2020)}]{3d-ssp-explain}
\bibinfo{author}{P.~Herscovitch},
\newblock \bibinfo{title}{A pioneering paper that provided a tool for accurate, observer-independent analysis of 18 f-fdg brain scans in neurodegenerative dementias},
\newblock \bibinfo{journal}{Journal of Nuclear Medicine} \bibinfo{volume}{61} (\bibinfo{year}{2020}) \bibinfo{pages}{140S--141S}. \DOIprefix\doi{10.2967/jnumed.120.252510}.

\bibitem[{Chen et~al.(2025)Chen, Su, Dumitrascu, Chen, Weidman, Caselli, Ashton, Reiman, and Wang}]{chen2025plasma}
\bibinfo{author}{Y.~Chen}, \bibinfo{author}{Y.~Su}, \bibinfo{author}{C.~Dumitrascu}, \bibinfo{author}{K.~Chen}, \bibinfo{author}{D.~Weidman}, 
\bibinfo{author}{R.~J.~Caselli}, \bibinfo{author}{N.~Ashton}, \bibinfo{author}{E.~M.~Reiman}, \bibinfo{author}{Y.~Wang},
\newblock \bibinfo{title}{Plasma-CycleGAN: Plasma biomarker-guided {MRI} to {PET} cross-modality translation using conditional CycleGAN},
\newblock in: \bibinfo{booktitle}{Proc. IEEE 22nd Int. Symp. Biomed. Imaging (ISBI)}, \bibinfo{pages}{1--5}, \bibinfo{year}{2025}.

\bibitem[{Chen et~al.(2025)Chen, Weng, Huang, Zhang, Dening, Hosseini, and Xiao}]{chen2025multi}
\bibinfo{author}{K.~Chen}, \bibinfo{author}{Y.~Weng}, \bibinfo{author}{Y.~Huang}, \bibinfo{author}{Y.~Zhang}, \bibinfo{author}{T.~Dening},
\bibinfo{author}{A.~A.~Hosseini}, \bibinfo{author}{W.~Xiao},
\newblock \bibinfo{title}{A multi-view learning approach with diffusion model to synthesize {FDG} {PET} from {MRI} {T1WI} for diagnosis of {Alzheimer's} disease},
\newblock \bibinfo{journal}{Alzheimer's \& Dementia} \bibinfo{volume}{21}~(\bibinfo{number}{2}) (\bibinfo{year}{2025}) \bibinfo{pages}{e14421}.

\bibitem[{Zotova et~al.(2025)Zotova, Pinon, Trombetta, Bouet, Jung, and Lartizien}]{zotova2025gan}
\bibinfo{author}{D.~Zotova}, \bibinfo{author}{N.~Pinon}, \bibinfo{author}{R.~Trombetta}, \bibinfo{author}{R.~Bouet}, \bibinfo{author}{J.~Jung}, \bibinfo{author}{C.~Lartizien},
\newblock \bibinfo{title}{{GAN}-based synthetic {FDG} {PET} images from {T1} brain {MRI} can serve to improve performance of deep unsupervised anomaly detection models},
\newblock \bibinfo{journal}{Comput. Methods Programs Biomed.} \bibinfo{volume}{265} (\bibinfo{year}{2025}) \bibinfo{pages}{108727}.

\bibitem[{Lin et~al.(2021)Lin, Lin, Chen, Zhang, Gao, Huang, Tong, Du, and Alzheimer's Disease Neuroimaging Initiative}]{lin2021bidirectional}
\bibinfo{author}{W.~Lin}, \bibinfo{author}{W.~Lin}, \bibinfo{author}{G.~Chen}, \bibinfo{author}{H.~Zhang}, \bibinfo{author}{Q.~Gao}, 
\bibinfo{author}{Y.~Huang}, \bibinfo{author}{T.~Tong}, \bibinfo{author}{M.~Du}, \bibinfo{author}{Alzheimer’s Disease Neuroimaging Initiative},
\newblock \bibinfo{title}{Bidirectional mapping of brain {MRI} and {PET} with 3D reversible {GAN} for the diagnosis of {Alzheimer’s} disease},
\newblock \bibinfo{journal}{Front. Neurosci.} \bibinfo{volume}{15} (\bibinfo{year}{2021}) \bibinfo{pages}{646013}.

\bibitem[{Yu et~al.(2024)Yu, Wu, Yue, Bozoki, and Liu}]{yu2024functional}
\bibinfo{author}{M.~Yu}, \bibinfo{author}{M.~Wu}, \bibinfo{author}{L.~Yue}, \bibinfo{author}{A.~Bozoki}, \bibinfo{author}{M.~Liu},
\newblock \bibinfo{title}{Functional imaging constrained diffusion for brain {PET} synthesis from structural {MRI}},
\newblock \bibinfo{journal}{arXiv preprint arXiv:2405.02504} (\bibinfo{year}{2024}).

\bibitem[{Kong et~al.(2021)Kong, Lian, Huang, Hu, and Zhou}]{kong2021breaking}
\bibinfo{author}{L.~Kong}, \bibinfo{author}{C.~Lian}, \bibinfo{author}{D.~Huang}, \bibinfo{author}{Y.~Hu}, \bibinfo{author}{Q.~Zhou} \emph{et~al.},
\newblock \bibinfo{title}{Breaking the dilemma of medical image-to-image translation},
\newblock \bibinfo{journal}{Adv. Neural Inf. Process. Syst.} \bibinfo{volume}{34} (\bibinfo{year}{2021}) \bibinfo{pages}{1964--1978}.

\bibitem[{Kim and Park(2024)}]{kim2024adaptive}
\bibinfo{author}{J.~Kim}, \bibinfo{author}{H.~Park},
\newblock \bibinfo{title}{Adaptive latent diffusion model for 3D medical image to image translation: Multi-modal magnetic resonance imaging study},
\newblock in: \bibinfo{booktitle}{Proc. IEEE/CVF Winter Conf. Appl. Comput. Vis. (WACV)}, \bibinfo{pages}{7604--7613}, \bibinfo{year}{2024}.


\end{thebibliography}

\end{document}